\documentclass{article} 
\usepackage{iclr2027_conference,times}
\usepackage{booktabs}

\usepackage{tabularx}
\usepackage{array}
\usepackage{multirow}
\usepackage[table]{xcolor}
\usepackage{amssymb}

\usepackage[most]{tcolorbox}

\definecolor{suppbg}{RGB}{245,247,249}
\definecolor{suppborder}{RGB}{225,229,234}

\usepackage{algorithm}
\usepackage{algpseudocode}

\usepackage[most]{tcolorbox}
\definecolor{suppbg}{RGB}{247,249,252}
\definecolor{suppborder}{RGB}{180,190,205}

\usepackage{amsthm}

\usepackage{centernot}

\usepackage{booktabs,xcolor,pifont}

\usepackage{enumitem}
\usepackage{multirow}

\usepackage{amsmath,amsfonts,bm}

\def\eqref#1{equation~\ref{#1}}

\def\1{\bm{1}}

\DeclareMathAlphabet{\mathsfit}{\encodingdefault}{\sfdefault}{m}{sl}
\SetMathAlphabet{\mathsfit}{bold}{\encodingdefault}{\sfdefault}{bx}{n}

\usepackage[table]{xcolor}
\usepackage{booktabs}

\definecolor{bestgreen}{RGB}{114,236,79}
\definecolor{lightgreen}{RGB}{207,243,197}
\usepackage{graphicx}
\definecolor{BestRed}{RGB}{248,190,190}
\definecolor{SecondOrange}{RGB}{251,211,170}
\definecolor{ThirdYellow}{RGB}{255,239,170}

\newcommand{\best}[1]{\cellcolor{BestRed}\textcolor{black}{#1}}
\newcommand{\second}[1]{\cellcolor{SecondOrange}\textcolor{black}{#1}}
\newcommand{\third}[1]{\cellcolor{ThirdYellow}\textcolor{black}{#1}}

\usepackage{hyperref}
\usepackage{url}
\usepackage[table]{xcolor}
\usepackage{graphicx}
\usepackage{float}    
\usepackage{caption}
\usepackage{amsthm}

\usepackage{pifont}

\title{EffGS: Efficient and High-Fidelity Gaussian Splatting}

\author{
Changbai Li$^{1*}$ \quad
Shuo Yang$^{1*}$ \quad
Yichen Yang$^{1}$ \quad
Shuwei Shao$^{2}$ \quad
Huobin Tan$^{1\dagger}$ \\[0.6em]
{\normalfont $^{1}$Beihang University \qquad $^{2}$Nanyang Technological University}
}

\iclrfinalcopy 
\begin{document}

\maketitle

\begingroup
\renewcommand{\thefootnote}{\fnsymbol{footnote}}
\footnotetext[1]{Equal contribution.}
\footnotetext[2]{Corresponding author.}
\endgroup

\begin{figure}[!h]
    \centering
    \includegraphics[width=\linewidth]{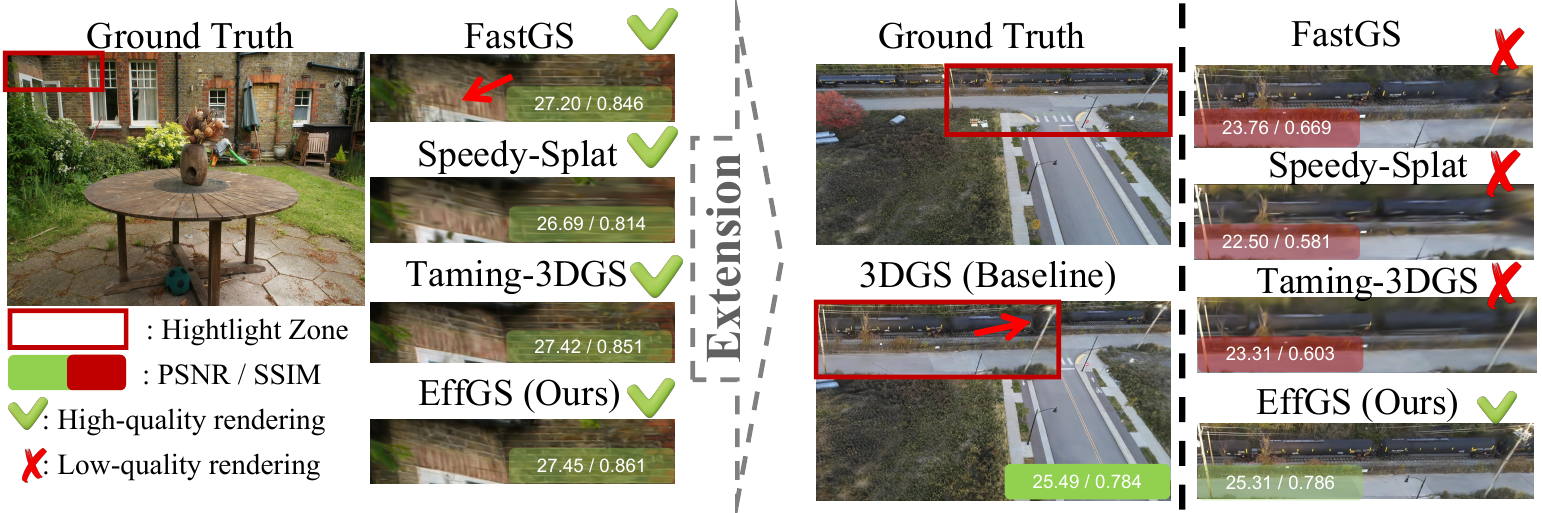} 
    \vspace{-2em}
    \caption{Compared with existing 3DGS acceleration methods, they can maintain high-quality rendering in small-scale scenes but suffer from degraded rendering quality when scaled up to city-level scenarios. In contrast, EffGS sustains high-fidelity rendering consistently, demonstrating superior scalability.}
    \label{fig:time1}
    \vspace{-1em}
\end{figure}

\begin{abstract}
3D Gaussian Splatting (3DGS) enables real-time novel view synthesis, but existing general-purpose acceleration methods suffer severe rendering quality degradation when extended to more complex, large-scale scenes. To address this issue, we propose EffGS, a more general acceleration framework that improves training and rendering efficiency while maintaining reconstruction quality comparable to or better than vanilla 3DGS across bounded and large-scale scenes. EffGS combines frequency-aware guidance, localized density control, and adaptive primitive scale modulation. First, an importance scoring mechanism combines pixel-wise reconstruction errors with a difference-of-Gaussians mask scheduled over training to provide stage-dependent spatial guidance. Second, localized densification and pruning restricts density modifications to Gaussians with valid projected footprints in the sampled views. Third, learnable per-Gaussian scale modulation adjusts effective primitive extent during optimization while retaining the Compact Box rasterization rule. Extensive experiments on bounded and large-scale scene datasets demonstrate a favorable balance between reconstruction quality, training time, and primitive count. Component ablations and matched-primitive-budget comparisons further support the effectiveness of the framework.
Code will be released at
 \url{https://github.com/CypressLi01/EffGS}.
\end{abstract}

\section{Introduction}

\begin{figure}[!t]

    \centering
    \includegraphics[width=\linewidth]{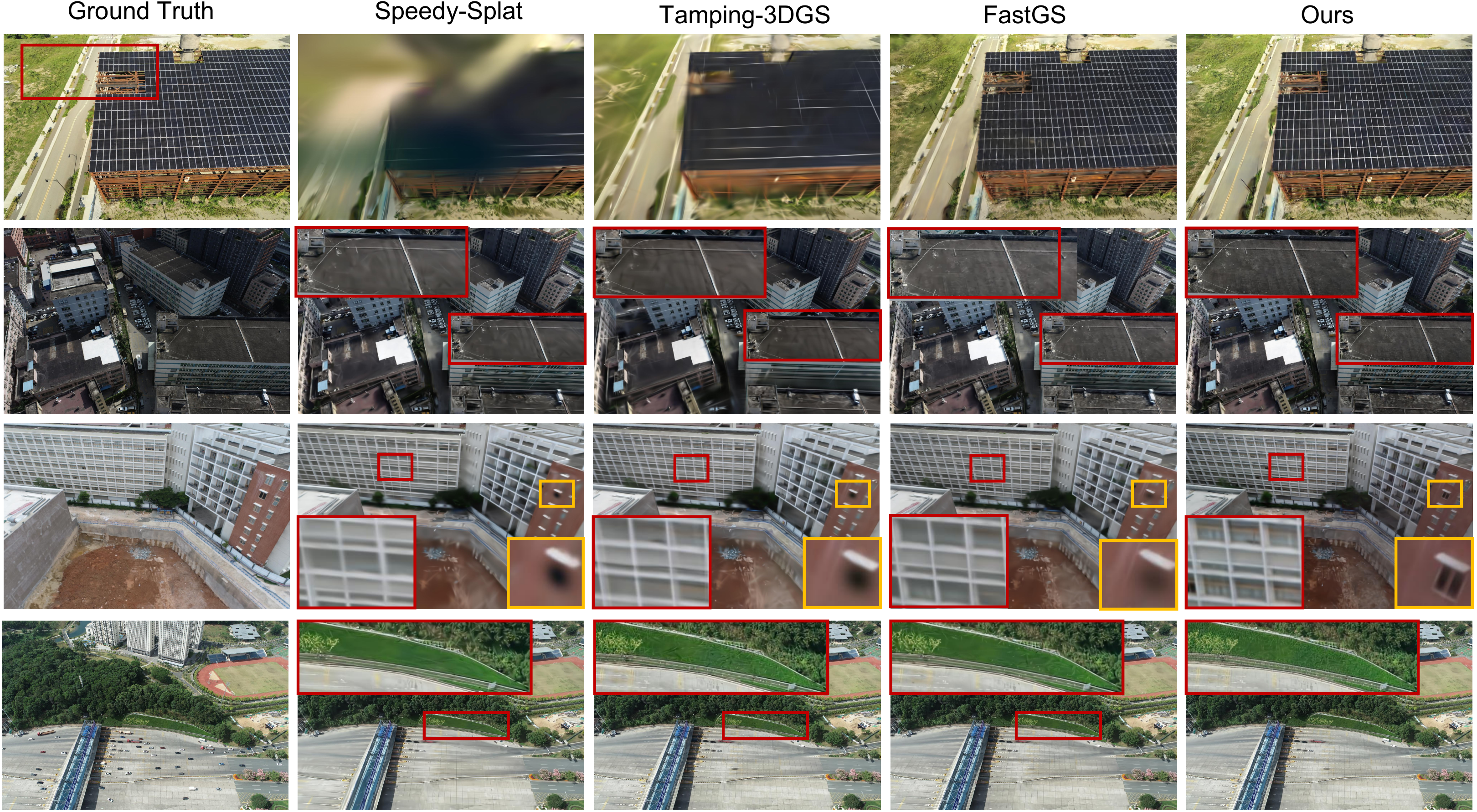} 
    \vspace{-2em}
    \caption{Qualitative results of ours and other methods in image rendering on Mill-19~\citep{mill}, Urbanscene3D~\citep{urbanscene} and GauU-Scene datasets~\citep{GauU-scene}.}
    \label{fig:render-city}
    \vspace{-1em}
\end{figure}

Neural Radiance Fields (NeRF)~\citep{nerf} advanced novel view
synthesis (NVS)~\citep{merf}, but their computational demands
motivated more efficient scene representations.
3D Gaussian Splatting (3DGS)~\citep{3dgs} uses explicit Gaussian
primitives and a tile-based rasterizer to achieve real-time,
photorealistic rendering.
However, adaptive density control can generate redundant
primitives, increasing optimization and rasterization
costs~\citep{lightgaussian}.
These costs become particularly important when reconstructing
large-scale environments.
Recent acceleration methods improve density control through
budget constraints~\citep{TimingGS, DashGaussian}, targeted
pruning~\citep{Speedy-Splat}, and multi-view
evaluation~\citep{fastgs}.
These approaches reduce computational cost, but preserving
rendering quality under aggressive primitive reduction remains
challenging.
Our city-scale comparisons
(Tab.~\ref{tab:large_scene_comparison} and
Fig.~\ref{fig:render-city}) illustrate this trade-off.
We focus on three aspects of the optimization pipeline:
the spatial information used to score primitives, the sampled
views supporting density-control decisions, and the effective
extent of each Gaussian.

\begin{figure}[!t]
    \centering
    \includegraphics[width=\linewidth]{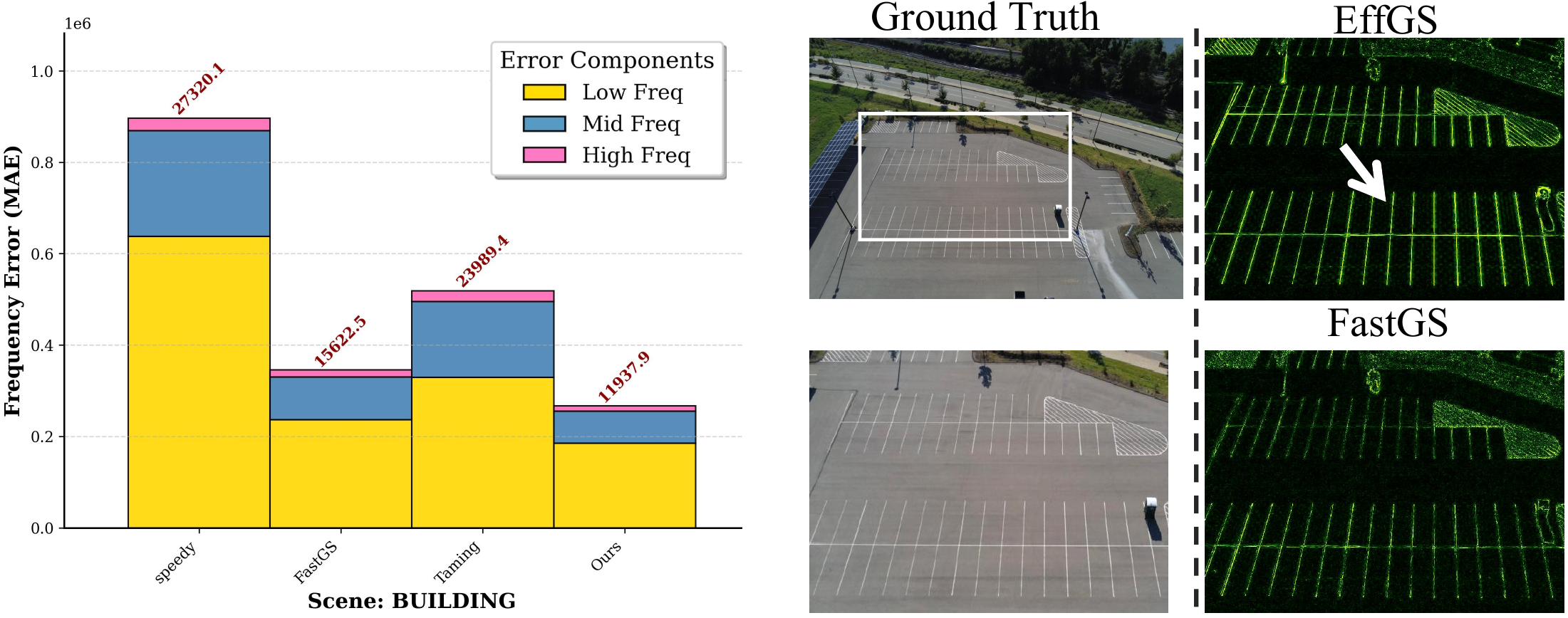}
    \vspace{-2em}
\caption{Spectral magnitude errors (left) and high-frequency discrepancy visualizations with ground-truth references (right) on Building~\citep{mill}; see Sec.~\ref{sec:frequency_aware_error_analysis}.}
    \vspace{-1em}
    \label{low-mid-higherror}
\end{figure}

1. \textbf{spatial detail should inform primitive allocation}.
Reconstruction errors identify regions that remain difficult
to render, while scale-dependent image responses provide
complementary information about local structure.
Combining these signals allows density control to emphasize
selected image structures at different training stages.
The spectral magnitude comparisons and spatial visualizations
in Fig.~\ref{low-mid-higherror} further motivate evaluating
detail preservation alongside standard reconstruction metrics.

2. \textbf{density-control decisions should account for
sampled-view coverage}.
In large-scale scenes, a small set of sampled cameras typically
covers only part of the Gaussian representation.
FastGS~\citep{fastgs} evaluates primitives using reconstruction
errors from randomly sampled views.
When applying view-based scores, primitives outside the sampled
coverage should be distinguished from those with valid projected
footprints.
Restricting density modifications to the latter localizes
the operation to the currently covered portion of the scene.
The FastGS+LDP comparison in Tab.~\ref{tab:fastgs+LDP}
supports the benefit of this restriction.

3. \textbf{primitive extent affects rasterization cost}.
Vanilla 3DGS~\citep{3dgs} uses a three-sigma extent to construct
rasterization bounds.
Speedy-Splat~\citep{Speedy-Splat} reduces redundant
Gaussian--tile pairs through precise tile intersection,
and FastGS~\citep{fastgs} further restricts support using
its Compact Box mechanism.
These culling rules operate on projected Gaussian geometry.
Adapting effective scales during optimization therefore
complements tile culling by changing the spatial support
to which the rule is applied.

We introduce \textbf{EffGS}, a framework that combines
frequency-aware scoring, localized density control, and
adaptive primitive compactness.
Our scoring mechanism couples pixel-wise reconstruction
errors with a scheduled difference-of-Gaussians mask.
The mask provides two-stage, scale-dependent spatial guidance
for importance scoring and reconstruction supervision.
Localized Densification and Pruning (LDP) restricts density
operations to Gaussians with valid projected footprints in
the sampled views, excluding primitives outside the current
active set.
Finally, a learnable scalar $\gamma_i$ uniformly modulates
each Gaussian's three base-scale components.
Joint optimization of this factor and the base scales
provides an additional pathway for adapting effective
primitive extent while preserving the Compact Box
tile-culling rule.
Experiments on bounded and urban-scale datasets show that
EffGS combines high reconstruction quality with efficient
training and real-time rendering.
Component ablations support the contribution of each module,
while matched-primitive-budget comparisons evaluate the
framework under controlled Gaussian counts.
We additionally report band-wise spectral magnitude errors
and high-frequency response discrepancies to complement
standard rendering metrics
(Sec.~\ref{sec:frequency_aware_error_analysis}).
Our main contributions are summarized as follows:
\begin{itemize}
    \item \textbf{Frequency-Aware Optimization:}
    We combine reconstruction errors with scheduled
    difference-of-Gaussians masks to provide stage-dependent
    spatial guidance for primitive allocation and
    reconstruction supervision.

    \item \textbf{Localized Density Control (LDP):}
    We restrict densification and pruning to Gaussians
    with valid projected footprints in sampled views,
    improving reconstruction quality in the evaluated
    large-scale scenes.

    \item \textbf{Adaptive Primitive Compactness:}
    We introduce learnable per-Gaussian scale modulation
    that interacts with optimization and density control
    to improve the quality--efficiency trade-off while
    retaining the Compact Box rasterization rule.
\end{itemize}

\section{Related Work}
\textbf{Neural Rendering.}
Novel View Synthesis has attracted significant attention in computer vision and graphics~\citep{surf3r}. Neural Radiance Fields (NeRF)~\cite{nerf} implicitly parameterize scene information via a multi-layer perceptron network and achieve high-quality image synthesis through volume rendering, greatly advancing NVS~\citep{mega-nerf,switchnerf}. However, due to its slow rendering efficiency, 3D Gaussian Splatting (3DGS)~\citep{3dgs} was subsequently introduced. 3DGS explicitly represents scenes with Gaussian primitives and employs GPU-friendly rasterization with alpha compositing for image rendering, thereby significantly accelerating optimization and enabling real-time rendering. Subsequently, a series of methods based on 3DGS have been proposed to enhance rendering quality~\citep{mip360-splatting,HRGS,PGSR,gof}.
Although 3DGS~\citep{3dgs} achieves impressive rendering results, the redundancy of Gaussian primitives considerably reduces its efficiency~\citep{lightgaussian,HRGS,ges,3D-GS,3DGS-LM}, especially when extending to city-scale scenes where multi-GPU training is often used to improve training efficiency~\citep{UrbanGS,Citygs,citygs-x,citygsv2,lin2024vastgaussian}, leading to increased training time. To accelerate 3DGS, many recent works have made substantial efforts.

\textbf{3DGS Acceleration.}
Density control is a major direction for accelerating
3DGS~\citep{fastgs}.
For densification, recent
works~\citep{DashGaussian,TimingGS,color-cued,revisingden}
refine primitive allocation through mechanisms such as
Gaussian budgets~\citep{TimingGS} and resolution-guided
scheduling~\citep{DashGaussian}.
For pruning, methods reduce the representation using
importance estimates and selection
strategies~\citep{lightgaussian,HRGS,Mini-Splatting,Speedy-Splat,3D-GS},
including Gaussian attributes, sampling-based
reduction~\citep{Mini-Splatting}, and approximate second-order
information~\citep{3D-GS,Speedy-Splat}.
FastGS~\citep{fastgs} evaluates pixel-wise reconstruction
errors from randomly sampled views to guide densification
and pruning.
These methods offer different trade-offs between primitive
count, computational cost, and rendering quality.
EffGS builds on error-based scoring and Compact Box
rasterization by incorporating scale-dependent spatial
guidance, restricting density operations to the sampled-view
active set, and introducing learnable per-Gaussian scale
modulation.
The combined framework targets efficient training with
high reconstruction fidelity across scene scales.

\section{Method}

We first revisit the basics of 3DGS
(Sec.~\ref{sec:prelim}).
We then introduce frequency-aware importance scoring, which
combines pixel-wise reconstruction errors with scheduled
difference-of-Gaussians masks
(Sec.~\ref{sec:freq_importance}).
Localized densification and pruning restricts density
modifications to Gaussians with valid projected footprints
in the sampled views
(Sec.~\ref{sec:local_densification_pruning}).
Next, adaptive primitive compactness uses learnable
per-Gaussian scale modulation to adjust effective spatial
support during optimization
(Sec.~\ref{sec:adaptive_compactness}).
Finally, we describe the training objective that combines
photometric, structural, mask-weighted, and scale-factor
regularization terms
(Sec.~\ref{Optimization}).
Additional frequency-domain analysis, including spectral
decomposition and band-wise reconstruction-error measurements,
is provided in the supplementary material
(Sec.~\ref{sec:spectral_analysis}).

\begin{figure}[!t]
    \centering

    \includegraphics[width=\linewidth]{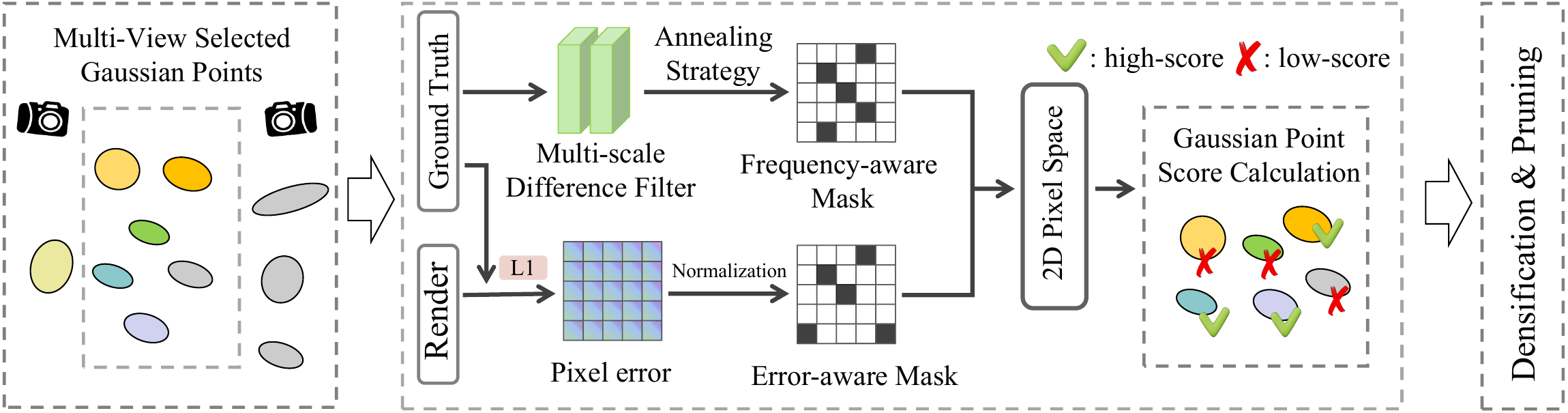} 

    \caption{\textbf{Overview of the proposed unified optimization pipeline.} Given a set of sampled views, we first identify the visible Gaussians. We then compute a unified 2D supervision signal by fusing a pixel-wise error-aware mask with a frequency-aware mask (extracted via an annealing schedule). Finally, we project the visible Gaussians to 2D to accumulate importance scores from the unified mask, guiding localized densification and pruning. }

    \label{pipeline}
\end{figure}

\subsection{Preliminary}
\label{sec:prelim}

We briefly review 3D Gaussian Splatting (3DGS)~\citep{3dgs}.
A scene is represented by a set of 3D Gaussians $\{G_k\}$.
Each Gaussian has position $\mathbf{p}_k$, opacity $o_k$, rotation
$R_k$, base-scale matrix
$S_k=\mathrm{diag}(s_k^x,s_k^y,s_k^z)$, and covariance
$\Sigma_k=R_kS_kS_k^\top R_k^\top$.
Its spherical-harmonic coefficients determine the
view-dependent RGB color $\mathbf{c}_k$.

For rendering, the covariance is projected to the image plane as
\begin{equation}
    \Sigma_k^{\mathrm{2D}}
    =
    J W \Sigma_k W^\top J^\top,
\end{equation}
where $W\in\mathbb{R}^{3\times3}$ is the rotational part of the
world-to-camera transform and $J$ is the projection Jacobian
evaluated at $\mathbf{p}_k$.
For image-plane coordinate $\mathbf{u}\in\mathbb{R}^2$, let
$\mathbf{p}_k^{\mathrm{2D}}$ denote the projected Gaussian center.
The rendered color is obtained by front-to-back alpha compositing:
\begin{equation}
    \hat{\mathbf{C}}(\mathbf{u})
    =
    \sum_k
    \mathbf{c}_k\alpha_k(\mathbf{u})
    \prod_{\ell<k}\left(1-\alpha_\ell(\mathbf{u})\right),
\end{equation}
\begin{equation}
    \alpha_k(\mathbf{u})
    =
    o_k
    \exp\!\left(
    -\frac{1}{2}
    (\mathbf{u}-\mathbf{p}_k^{\mathrm{2D}})^\top
    (\Sigma_k^{\mathrm{2D}})^{-1}
    (\mathbf{u}-\mathbf{p}_k^{\mathrm{2D}})
    \right).
\end{equation}

3DGS uses adaptive density control: Gaussians with large positional
gradients are cloned or split, while those with low opacity or excessive
scale are pruned. The standard objective is
\begin{equation}
    \mathcal{L}
    =
    (1-\lambda_{\mathrm{dssim}})\mathcal{L}_1
    +
    \lambda_{\mathrm{dssim}}
    (1-\mathcal{L}_{\mathrm{SSIM}}),
\end{equation}
where $\lambda_{\mathrm{dssim}}=0.2$.

\subsection{Frequency-Aware Importance Scoring}
\label{sec:freq_importance}

We combine reconstruction errors with scheduled image responses
to guide primitive allocation (Fig.~\ref{pipeline}).

\paragraph{Frequency mask generation.}
We convert each ground-truth image to grayscale and downsample it
to obtain $I_{\mathrm{down}}$. Let $\mathcal{G}_{\sigma}$ denote a
normalized Gaussian kernel with standard deviation $\sigma$. We compute
\begin{equation}
\label{eq:freq_response}
H
=
\left|
I_{\mathrm{down}} * \mathcal{G}_{\sigma_1}
-
I_{\mathrm{down}} * \mathcal{G}_{\sigma_2}
\right|,
\qquad
\sigma_2=2\sigma_1,
\end{equation}
where $*$ denotes convolution.
The bandwidth follows the piecewise schedule in
Sec.~\ref{sec:hf_mask}, controlled by
$f(t)=100t/T_{\mathrm{total}}$, where $t$ is the current training
iteration and $T_{\mathrm{total}}$ is the total number of iterations.
After upsampling and normalization, we threshold the complemented
response for $f<50$ and the direct response for $f\geq50$ to obtain
the frequency-aware mask $M_{\mathrm{freq}}$.
This provides stage-dependent spatial guidance, with the decreasing
bandwidth in the second stage emphasizing finer structures.

\paragraph{Composite error map.}
For sampled view $j$ with pixel grid $\mathcal{P}_j$, let
$I_{\mathrm{rend}}^j$ and $I_{\mathrm{gt}}^j$ denote the rendered and
ground-truth images, respectively, and let $M_{\mathrm{freq}}^j$ be
the corresponding frequency-aware mask.
The raw RGB reconstruction error is
\[
e^j(\mathbf{u})
=
\frac{1}{3}
\sum_{c=1}^{3}
\left|
I_{\mathrm{rend},c}^j(\mathbf{u})
-
I_{\mathrm{gt},c}^j(\mathbf{u})
\right|,
\qquad
\bar e^j
=
\mathcal{N}_{\mathcal{P}_j}[e^j],
\]
where $\mathcal{N}_{\mathcal{S}}$ denotes min--max normalization
over index set $\mathcal{S}$, with constant inputs mapped to zero.
The composite mask is
\begin{equation}
\label{eq:error_mask}
M_{\mathrm{err}}^j(\mathbf{u})
=
\mathbb{I}\!\left[
\bar e^j(\mathbf{u})>\tau
\;\lor\;
\left(
M_{\mathrm{freq}}^j(\mathbf{u})=1
\;\land\;
\bar e^j(\mathbf{u})>0.5\tau
\right)
\right],
\end{equation}
where $\mathbb{I}[\cdot]$ is the indicator function and
$\tau\in(0,1)$ is the normalized error threshold.

\paragraph{Densification scores.}
Let $\mathcal{V}_i$ contain the sampled views in which Gaussian $i$
has a valid projected footprint, and let $\mathcal{A}_t$ denote the
active primitive set at the current density update
(Sec.~\ref{sec:local_densification_pruning}).
For each view, $\Omega_i^j$ contains the pixels at which Gaussian $i$
is processed and contributes during compositing. Define
\[
n_i^j
=
\sum_{\mathbf{u}\in\Omega_i^j}
M_{\mathrm{err}}^j(\mathbf{u}).
\]
The valid-view average and densification score are
\begin{equation}
\label{eq:base_count}
C_i
=
\frac{1}{|\mathcal{V}_i|}
\sum_{j\in\mathcal{V}_i}n_i^j,
\qquad i\in\mathcal{A}_t,
\end{equation}
\begin{equation}
\label{eq:densification_score}
s_d^i(t)=\omega(t)C_i,
\end{equation}
where $\omega(t)$ is a positive, nondecreasing clipped-linear
schedule over the densification interval. Exact normalization,
pixel-support, and scheduling definitions are provided in
Sec.~\ref{sec:density_control_details}.

\paragraph{Pruning scores.}
Following FastGS~\citep{fastgs}, we weight each per-view count by
the corresponding image-level photometric loss:
\[
E_{\mathrm{photo}}^j
=
(1-\lambda_{\mathrm{dssim}})\mathcal{L}_1^j
+
\lambda_{\mathrm{dssim}}
\left(
1-\operatorname{SSIM}
(I_{\mathrm{rend}}^j,I_{\mathrm{gt}}^j)
\right),
\qquad
\mathcal{L}_1^j
=
\frac{1}{|\mathcal{P}_j|}
\sum_{\mathbf{u}\in\mathcal{P}_j}e^j(\mathbf{u}).
\]
The pruning score is
\begin{equation}
\label{eq:pruning_score}
Q_i
=
\sum_{j\in\mathcal{V}_i}
n_i^jE_{\mathrm{photo}}^j,
\qquad
s_p^i
=
\mathcal{N}_{\mathcal{A}_t}[Q](i).
\end{equation}
Thus, densification uses a valid-view average, whereas pruning
retains summed, loss-weighted counts.
$E_{\mathrm{photo}}^j$ is used only for scoring; parameter
optimization follows Sec.~\ref{Optimization}.

\begin{figure*}[t]
  \centering
  \centering \includegraphics[width=1.0\textwidth]{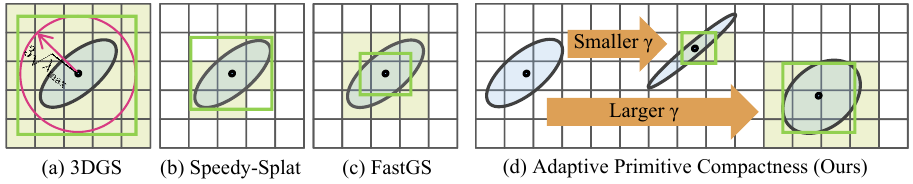}
   \caption{\textbf{Learnable primitive compactness.} Compared with vanilla 3DGS~\citep{3dgs}, Speedy-Splat~\citep{Speedy-Splat}, and FastGS~\citep{fastgs}, our method learns per-Gaussian compactness, reducing redundant pairs and improving fidelity.} 
   \label{fig:main}
\vspace{-0.2cm}
\end{figure*}

\begin{figure}[!t]
    \label{fig:render-mip-tnt}
    \centering
    \includegraphics[width=\linewidth]{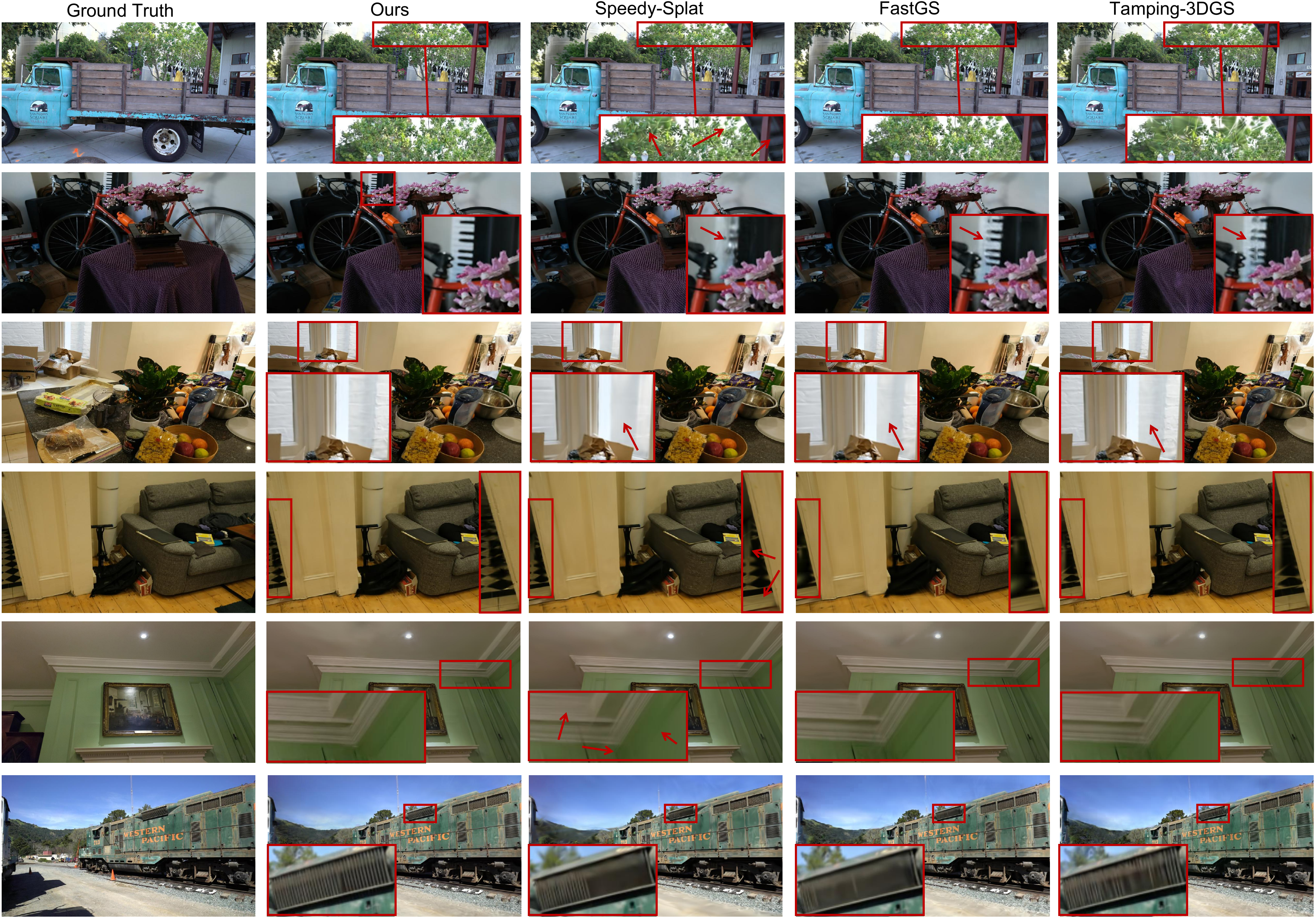} 
    \vspace{-2em}
    \caption{Qualitative results of ours and other methods in image rendering on Deep Blending~\citep{deep}, Mip-NeRF 360~\citep{Mip-NeRF360} and Tanks \& Temples datasets~\citep{TNT}.}
\end{figure}

\subsection{View-Visible Local Densification and Pruning}
\label{sec:local_densification_pruning}

We restrict density modifications to the portion of the representation
covered by the sampled views.

\paragraph{Active primitives.}
At each scheduled update, we sample $K$ distinct training views
uniformly without replacement.
Let $\mathcal{G}_t$ denote the index set of Gaussian primitives before
the update, and let $r_i^j$ be the projected radius returned by the
rasterizer. We define
\begin{equation}
\label{eq:visibility}
\mathcal{V}_i
=
\{j:r_i^j>0\},
\qquad
|\mathcal{V}_i|
=
\sum_{j=1}^{K}\mathbb{I}[r_i^j>0],
\qquad
\mathcal{A}_t
=
\{i\in\mathcal{G}_t:|\mathcal{V}_i|>0\}.
\end{equation}
This criterion defines projected-footprint eligibility rather than
explicit occlusion visibility; actual counted contributions are
determined by $\Omega_i^j$. Density control is skipped when
$\mathcal{A}_t=\varnothing$.

\paragraph{Pruning.}
At a pruning step, mandatory removals are
\[
\mathcal{H}_t
=
\left\{
i\in\mathcal{A}_t:
o_i<0.1
\;\lor\;
\gamma_i<0.01
\right\},
\]
where $\gamma_i$ is the learnable scale-modulation factor introduced
in Sec.~\ref{sec:adaptive_compactness}.
The remaining candidates are
\[
\mathcal{P}_t
=
\left\{
i\in\mathcal{A}_t\setminus\mathcal{H}_t:
s_p^i>\tau_p
\right\},
\qquad
0\leq\tau_p<1.
\]
We sample
$b_t=\lfloor\rho|\mathcal{P}_t|\rfloor$
candidates without replacement using weights proportional to
$s_p^i$, where $\rho\in[0,1]$.
The removal set $\mathcal{R}_t$ is their union with
$\mathcal{H}_t$.

\paragraph{Densification and execution.}
At a densification step, surviving active primitives satisfying
$s_d^i(t)>\tau_d$ are selected, where $\tau_d\geq0$ is the
densification-score threshold.
A primitive is cloned and shifted along its positional gradient if
its largest base-scale component is below $\tau_s>0$; otherwise, it
is replaced by two smaller Gaussians.
New primitives inherit the parent's unconstrained modulation parameter
$\beta_i\in\mathbb{R}$, with
$\gamma_i=2\operatorname{sigmoid}(\beta_i)\in(0,2)$, while their
remaining attributes follow the underlying clone/split updates.
All scores are computed from the same pre-update representation.
When pruning and densification coincide, pruning is applied first;
new primitives are scored only at the next density update.
Primitives outside $\mathcal{A}_t$ undergo neither operation.
The complete execution protocol is given in
Sec.~\ref{sec:density_control_details}.

\subsection{Adaptive Primitive Compactness}
\label{sec:adaptive_compactness}

During rasterization, the projected extent of each Gaussian determines
the tiles that require processing.
Vanilla 3DGS~\citep{3dgs} uses a three-sigma extent, while
Speedy-Splat~\citep{Speedy-Splat} and FastGS~\citep{fastgs} reduce
redundant Gaussian--tile evaluations through more restrictive
culling rules.
We retain the Compact Box mechanism of FastGS and introduce a
learnable per-Gaussian scale modulation.
For Gaussian $i$,
\begin{equation}
\label{eq:scale_modulation}
\widetilde{S}_i
=
\gamma_iS_i,
\qquad
\widetilde{\Sigma}_i
=
\gamma_i^2\Sigma_i .
\end{equation}
With the base scales fixed, $\gamma_i$ changes the overall spatial
extent while preserving anisotropic axis ratios.
Because both $S_i$ and $\gamma_i$ remain learnable, the modulation
acts as an additional optimization and density-control variable rather
than enlarging the family of representable covariances.
It therefore provides an additional pathway for adapting effective
primitive support while retaining the Compact Box tile-culling rule.
Further details are provided in Sec.~\ref{sec:compact_gaussian}.

\begin{table*}[t]
\centering
\caption{
Quantitative comparison on Mip-NeRF 360~\citep{Mip-NeRF360},
Deep Blending~\citep{blender}, and Tanks \& Temples~\citep{TNT}.
}
\label{tab:comparison}
 
\setlength{\tabcolsep}{1.2pt}
\resizebox{\textwidth}{!}{
\begin{tabular}{l cccccc cccccc cccccc}
\toprule
\multirow{2}{*}{Method}
& \multicolumn{6}{c}{Mip-NeRF 360}
& \multicolumn{6}{c}{Deep Blending}
& \multicolumn{6}{c}{Tanks \& Temples} \\
\cmidrule(lr){2-7}
\cmidrule(lr){8-13}
\cmidrule(lr){14-19}
& Time $\downarrow$
& PSNR $\uparrow$
& SSIM $\uparrow$
& LPIPS $\downarrow$
& $N_{GS}$ $\downarrow$
& FPS $\uparrow$
& Time $\downarrow$
& PSNR $\uparrow$
& SSIM $\uparrow$
& LPIPS $\downarrow$
& $N_{GS}$ $\downarrow$
& FPS $\uparrow$
& Time $\downarrow$
& PSNR $\uparrow$
& SSIM $\uparrow$
& LPIPS $\downarrow$
& $N_{GS}$ $\downarrow$
& FPS $\uparrow$ \\
\midrule

3DGS
& 20.93 & 27.53 & 0.812 & 0.221 & 2.63M & 146
& 19.77 & 29.71 & 0.903 & \second{0.241} & 2.46M & 158
& 11.34 & 23.71 & \third{0.850} & \second{0.170} & 1.57M & 195 \\

3DGS-LM
& 10.37 & 27.42 & 0.810 & 0.222 & 3.36M & 220
& 9.83 & 29.61 & 0.905 & 0.248 & 2.71M & 248
& 6.15 & 23.53 & 0.842 & 0.184 & 1.81M & 289 \\

Mini-Splatting
& 14.67 & 27.32 & \second{0.821} & \second{0.217} & 0.53M & \second{567}
& 13.35 & \third{29.99} & \second{0.907} & \third{0.244} & 0.56M & 624
& 9.06 & 23.46 & 0.844 & 0.181 & \third{0.30M} & \best{756} \\

Speedy-Splat
& 13.38 & 26.91 & 0.781 & 0.295 & \best{0.30M} & \third{552}
& 10.75 & 29.42 & 0.898 & 0.272 & \second{0.25M} & \second{664}
& 6.32 & 23.38 & 0.816 & 0.242 & \best{0.18M} & \second{691} \\

Taming-3DGS
& \third{5.14} & 27.48 & 0.794 & 0.261 & 0.68M & 221
& \third{3.06} & 29.50 & 0.894 & 0.278 & \third{0.29M} & 352
& \third{2.71} & 23.89 & 0.833 & 0.214 & 0.32M & 379 \\

DashGaussian
& 6.35 & \second{27.73} & \third{0.817} & \third{0.218} & 2.40M & 155
& 4.16 & 29.65 & \third{0.906} & 0.246 & 1.94M & 208
& 4.28 & \third{24.00} & \second{0.853} & \third{0.178} & 1.21M & 240 \\

FastGS
& \best{1.93} & \third{27.56} & 0.797 & 0.261 & \second{0.40M} & \best{579}
& \best{1.28} & \best{30.03} & 0.901 & 0.270 & \best{0.22M} & \best{714}
& \best{1.32} & \second{24.15} & 0.839 & 0.210 & \second{0.24M} & \third{655} \\

Ours
& \second{3.07} & \best{27.77} & \best{0.826} & \best{0.207} & \third{0.48M} & 478
& \second{2.94} & \second{30.01} & \best{0.910} & \best{0.238} & \second{0.25M} & \third{632}
& \second{1.94} & \best{24.22} & \best{0.861} & \best{0.166} & 0.44M & 572 \\

\bottomrule
\end{tabular}
}
\end{table*}

\subsection{Optimization}
\label{Optimization}

We further use $M_{\mathrm{freq}}$ to spatially weight reconstruction
errors during training.
For the currently sampled view, let $\mathcal{P}_{\mathrm{img}}$
denote its pixel grid and $C_{\mathrm{rgb}}=3$.
Using $I_{\mathrm{rend}}$ and $I_{\mathrm{gt}}$ for the rendered and
ground-truth images,
\[
\mathcal{L}_1
=
\frac{1}
{C_{\mathrm{rgb}}|\mathcal{P}_{\mathrm{img}}|}
\sum_{\mathbf{u}\in\mathcal{P}_{\mathrm{img}}}
\sum_{c=1}^{C_{\mathrm{rgb}}}
\left|
I_{\mathrm{rend},c}(\mathbf{u})
-
I_{\mathrm{gt},c}(\mathbf{u})
\right|.
\]
The frequency-guided loss is
\begin{equation}
\label{eq:freq_loss}
\mathcal{L}_{\mathrm{freq}}
=
\frac{1}
{C_{\mathrm{rgb}}|\mathcal{P}_{\mathrm{img}}|}
\sum_{\mathbf{u}\in\mathcal{P}_{\mathrm{img}}}
\sum_{c=1}^{C_{\mathrm{rgb}}}
M_{\mathrm{freq}}(\mathbf{u})
\left|
I_{\mathrm{rend},c}(\mathbf{u})
-
I_{\mathrm{gt},c}(\mathbf{u})
\right|.
\end{equation}
The mask is shared across color channels and uses the full-image
denominator, so an empty mask yields zero masked loss without a
separate normalization rule.
Let
$\mathcal{L}_{\mathrm{SSIM}}
=
\operatorname{SSIM}(I_{\mathrm{rend}},I_{\mathrm{gt}})$.
The complete objective is
\begin{equation}
\label{eq:total_loss}
\begin{aligned}
\mathcal{L}
={}&
(1-\lambda_{\mathrm{dssim}}-\lambda_{\mathrm{freq}})
\mathcal{L}_1
+
\lambda_{\mathrm{dssim}}
(1-\mathcal{L}_{\mathrm{SSIM}})
\\
&+
\lambda_{\mathrm{freq}}
\mathcal{L}_{\mathrm{freq}}
+
\mathcal{R}_{\gamma},
\end{aligned}
\end{equation}
where $\lambda_{\mathrm{dssim}}=0.2$ and
$\lambda_{\mathrm{freq}}=0.1$.
Let $\mathcal{G}$ denote the current primitive index set and
$N=|\mathcal{G}|$. The scale-factor regularizer is
\[
\mathcal{R}_{\gamma}
=
\begin{cases}
\displaystyle
\frac{\lambda_{\gamma}}{N}
\sum_{i\in\mathcal{G}}\gamma_i^2,
& N>0,\\[3pt]
0,&N=0,
\end{cases}
\qquad
\lambda_{\gamma}>0.
\]
The mean reduction prevents its scale from increasing solely with
the primitive count. The regularizer introduces a shrinkage preference
on the modulation factors, but is not a direct penalty on effective
Gaussian volume.

\begin{table*}[t]
\centering
\caption{
Quantitative comparison of the averaged metrics on
Mill-19~\citep{mill}, UrbanScene3D~\citep{urbanscene}, and
GauU-Scene~\citep{GauU-scene}.
}
\label{tab:large_scene_comparison}

\setlength{\tabcolsep}{1.5pt}
\resizebox{\textwidth}{!}{
\begin{tabular}{l cccccc cccccc cccccc}
\toprule
\multirow{2}{*}{Method}
& \multicolumn{6}{c}{Mill-19}
& \multicolumn{6}{c}{UrbanScene3D}
& \multicolumn{6}{c}{GauU-Scene} \\
\cmidrule(lr){2-7}
\cmidrule(lr){8-13}
\cmidrule(lr){14-19}
& Time (m) $\downarrow$
& SSIM $\uparrow$
& PSNR $\uparrow$
& LPIPS $\downarrow$
& $N_{GS}$ $\downarrow$
& FPS $\uparrow$
& Time (m) $\downarrow$
& SSIM $\uparrow$
& PSNR $\uparrow$
& LPIPS $\downarrow$
& $N_{GS}$ $\downarrow$
& FPS $\uparrow$
& Time (m) $\downarrow$
& SSIM $\uparrow$
& PSNR $\uparrow$
& LPIPS $\downarrow$
& $N_{GS}$ $\downarrow$
& FPS $\uparrow$ \\
\midrule

3DGS
& 196
& \second{0.735}
& \second{23.06}
& \second{0.292}
& 10.76
& $<50$
& 180
& 0.763
& \best{21.69}
& \best{0.252}
& 6.12
& $<50$
& 176
& \second{0.736}
& \third{23.66}
& \best{0.268}
& 6.58
& $<50$ \\

PGSR
& 212
& 0.603
& 20.12
& 0.447
& 12.41
& $<50$
& 177
& \best{0.780}
& 20.01
& 0.285
& 5.87
& $<50$
& 187
& 0.593
& 20.57
& 0.421
& 5.98
& $<50$ \\

Mip-Splatting
& 187
& \third{0.703}
& \third{22.35}
& \third{0.334}
& 18.44
& $<50$
& 192
& \third{0.775}
& \third{21.37}
& \second{0.266}
& 13.07
& $<50$
& 184
& 0.660
& 21.40
& 0.341
& 14.50
& $<50$ \\

Taming-3DGS
& \third{21}
& 0.551
& 21.22
& 0.510
& \second{0.75}
& 88
& \third{23}
& 0.633
& 19.74
& 0.460
& \best{0.48}
& 73
& \third{29}
& \third{0.714}
& \second{23.77}
& \third{0.308}
& 5.92
& 84 \\

Speedy-Splat
& 45
& 0.522
& 19.70
& 0.547
& \best{0.54}
& \second{204}
& 54
& 0.658
& 19.52
& 0.432
& \second{0.56}
& \second{181}
& 45
& 0.646
& 22.42
& 0.401
& \best{0.73}
& \second{185} \\

FastGS
& \best{12}
& 0.662
& 22.32
& 0.391
& \third{1.60}
& \best{251}
& \best{11}
& 0.708
& 20.24
& 0.351
& \third{1.28}
& \best{228}
& \best{13}
& 0.707
& 23.47
& 0.329
& \second{2.34}
& \best{236} \\

EffGS
& \second{20}
& \best{0.755}
& \best{23.68}
& \best{0.279}
& 3.76
& \third{145}
& \second{22}
& \second{0.778}
& \second{21.50}
& \third{0.281}
& 3.05
& \third{148}
& \second{21}
& \best{0.756}
& \best{24.23}
& \second{0.273}
& \third{4.23}
& \third{124} \\

\bottomrule
\end{tabular}
}
\end{table*}

\section{Experiment}
We first describe the datasets and implementation details in
Sec.~\ref{sec:Experimental Setup}. We then evaluate EffGS on both
bounded and city-scale scenes and compare it with representative
Gaussian Splatting methods in Sec.~\ref{sec:results_analysis}.
Finally, we analyze the contribution of each component through
ablation studies in Sec.~\ref{sec:ablation}.

\subsection{Experimental Setup}
\label{sec:Experimental Setup}

\textbf{{Datasets.} }
To comprehensively evaluate the rendering fidelity, computational efficiency, and scalability of our proposed method, we conduct experiments across a diverse range of scene configurations. For object-centric and bounded indoor/outdoor environments, we utilize the Mip-NeRF 360~\citep{Mip-NeRF360}, Deep-Blending~\citep{deep}, and Tanks and Temples~\citep{TNT} datasets. Furthermore, to rigorously assess the performance and robustness of our method in large-scale, complex, and unbounded scenarios, we evaluate it on the Mill-19~\citep{mill}, UrbanScene3D~\citep{urbanscene}, and GauU-Scene~\citep{GauU-scene} datasets.

\textbf{{Implementation Details.}}
EffGS and all baselines are implemented in PyTorch.
EffGS, Speedy-Splat~\citep{Speedy-Splat}, and
FastGS~\citep{fastgs} are evaluated on identical 24GB GPUs,
while baseline configurations that exceed this memory capacity on
large-scale scenes are evaluated using NVIDIA A800 GPUs.
Additional implementation details and extended experiments are
provided in the supplementary material (see Sec.~\ref{sec:exp_details}).

\subsection{Results Analysis}
\label{sec:results_analysis}
In this section, we first compare EffGS with representative Gaussian
Splatting methods on bounded scenes, and then evaluate its scalability
against efficient large-scene reconstruction methods on city-scale
datasets.
In the supplementary material, we further extend EffGS to multi-GPU
training and compare it with several city-scale multi-GPU reconstruction
methods, together with a detailed analysis of GPU memory consumption.
See Sec.~\ref{More experiments} for more details.

\textbf{{Performance on Bounded Scenes.}}
EffGS achieves a favorable balance between rendering quality and
training efficiency (Tab.~\ref{tab:comparison}).
Among the methods compared in this table, it achieves the highest
PSNR on Mip-NeRF 360 and Tanks \& Temples, reaching 27.77 and
24.22, respectively.
Across all three benchmarks, EffGS improves PSNR, SSIM, and LPIPS
over vanilla 3DGS~\citep{3dgs} while substantially reducing training
time.
Compared with FastGS, it achieves higher SSIM and lower LPIPS
on all three datasets, with additional training cost.
These results demonstrate the quality--efficiency benefits
of the complete framework.

\textbf{{Scalability to City-Scale Scenes.}}
As shown in Tab.~\ref{tab:large_scene_comparison}, EffGS
outperforms Taming-3DGS, Speedy-Splat, and FastGS in PSNR,
SSIM, and LPIPS across all three city-scale datasets.
Among the methods in this table, it achieves the best PSNR
and SSIM on Mill-19 and GauU-Scene, while maintaining
competitive quality on UrbanScene3D.
With training times of 20--22 minutes and rendering speeds
of 124--148 FPS, EffGS combines efficient training with
real-time rendering in large-scale scenes.

\begin{table}[t]
    \centering
    \caption{Component ablations on GauU-Scene~\citep{GauU-scene},
    evaluating reconstruction quality and computational cost.}
    \label{tab:ablation}

    \resizebox{\linewidth}{!}{
    \begin{tabular}{lccccccc}
        \toprule
        \multirow{2}{*}{\textbf{Ablation Item}} & \multicolumn{3}{c}{\textbf{Rendering Quality}} & \multicolumn{4}{c}{\textbf{Efficiency Metrics}} \\
        \cmidrule(lr){2-4} \cmidrule(lr){5-8}
        & SSIM $\uparrow$ & PSNR $\uparrow$ & LPIPS $\downarrow$ & Time (Min) $\downarrow$ & Size (GB) $\downarrow$ & GS (M) $\downarrow$ & Mem (G) $\downarrow$ \\
        \midrule
        (a) w/o Frequency-aware Mask & 0.752 & 24.12 & 0.278 & 22 & 1.15 & 4.51 & 15.6 \\
        (b) w/o Error-aware Mask & 0.748 & 23.98 & 0.284 & 21 & 1.23 & 4.42 & 14.8 \\
        (c) w/o Learnable Compactness ($\gamma_i$) & 0.745 & 23.85 & 0.288 & 28 & 1.25 & 4.98 & 16.5 \\
        (d) w/o Local Densification and Pruning (LDP) & 0.741 & 23.75 & 0.295 & \textbf{19} & \textbf{0.98} & \textbf{3.85} & \textbf{14.2} \\
        (e) w/o Frequency-aware Loss ($\mathcal{L}_{freq}$) & 0.753 & 24.16 & 0.276 & \textbf{19} & 1.05 & 4.20 & 15.3 \\
        \midrule
        \textbf{Full (Ours)} & \textbf{0.756} & \textbf{24.23} & \textbf{0.273} & 21 & 1.07 & 4.23 & 14.8 \\
        \bottomrule
    \end{tabular}
    }
\end{table}

\subsection{Ablation Study}
\label{sec:ablation}

We evaluate the contribution of each component on the GauU-Scene
dataset (Tab.~\ref{tab:ablation}).
Additional measurements of the frequency extraction module's
memory usage are provided in Sec.~\ref{sec:memory_freq}.
We further conduct matched-Gaussian-budget comparisons on both
bounded and large-scale scenes to control for differences in
primitive count (Tabs.~\ref{tab:supp_reference_and_matched}
and~\ref{tab:supp_iso_budget_large_scale}).

\textbf{{Frequency and Error-Aware Guidance.}}
Removing the frequency-aware mask (row a), the error-aware mask
(row b), or the frequency-guided loss $\mathcal{L}_{\mathrm{freq}}$
(row e) reduces PSNR and SSIM and increases LPIPS.
These results support combining reconstruction errors with
scale-dependent spatial guidance for density control and
reconstruction supervision.

\textbf{{Adaptive Learnable Compactness ($\gamma_i$).}}
Removing the learnable scale-modulation component (row c)
increases the Gaussian count to 4.98M, training time to
28 minutes, and peak memory usage to 16.5GB, while reducing
PSNR to 23.85.
These results support the contribution of scale reparameterization
and its interaction with density control to both reconstruction
quality and efficiency.
The extended evaluation in Tab.~\ref{tab:supp_compactness_extended}
further confirms this trend across representative bounded and
large-scale scenes, where learnable compactness consistently reduces
frequency-specific reconstruction errors while improving rendering
quality with fewer Gaussian primitives.

\textbf{{Local Densification and Pruning (LDP).}}
Removing LDP (row d) reduces the Gaussian count, training time,
and memory usage, but yields the lowest reconstruction quality
among the evaluated variants.
Retaining LDP improves all three quality metrics by restricting
density modifications to primitives with valid projected
footprints in the sampled views.
The results support this local restriction as a useful component
of the framework's quality--efficiency trade-off.
As shown in Tab.~\ref{tab:fastgs+LDP}, adding LDP to FastGS
consistently improves reconstruction quality on the Russian scene,
increasing SSIM from 0.741 to 0.758 and PSNR from 23.48 to 23.89,
while reducing LPIPS from 0.294 to 0.257.

\section{Conclusion}

We presented EffGS, a framework for efficient, high-fidelity
Gaussian splatting across bounded and city-scale scenes.
Frequency-aware scoring combines reconstruction errors with
scale-dependent spatial masks, while localized density
control restricts densification and pruning to primitives
with valid projected footprints in sampled views.
Learnable per-Gaussian scale modulation provides an additional
optimization pathway for adapting effective spatial support.
Together, these components achieve a favorable balance
between reconstruction quality, training time, and
primitive count.
Component ablations and matched-primitive-budget comparisons
support the effectiveness of the framework, with spectral
magnitude analysis complementing standard rendering metrics.

\section*{AI Use Statement}
Generative AI tools were used only for language polishing, grammar correction, and improving the clarity of the manuscript. They were not used to generate experimental results, perform data analysis, or make scientific claims. All technical content, experiments, and conclusions were produced and verified by the authors.

\section*{Ethics Statement}
This work focuses on efficient 3D scene reconstruction and does not involve human subjects, personal data, or sensitive attributes. All experiments are conducted on publicly available research datasets following their intended academic use. We are not aware of any direct ethical concerns specific to the proposed method.

\section*{Reproducibility Statement}
We provide detailed descriptions of the model architecture, optimization objectives, training schedule, hyperparameters, datasets, and evaluation protocols in the main paper and supplementary material. All experiments use consistent settings unless otherwise specified. We plan to release the implementation and configuration files to facilitate reproduction of the reported results.

\bibliography{iclr2027_conference}
\bibliographystyle{iclr2027_conference}

\newpage
\appendix

\begin{tcolorbox}[
    enhanced,
    breakable,
    colback=suppbg,
    colframe=suppborder,
    boxrule=0.4pt,
    arc=3mm,
    left=4mm,
    right=4mm,
    top=3mm,
    bottom=3mm,
    before skip=0pt,
    after skip=0pt
]

\begin{center}
{\Large\bfseries Supplementary Material}
\end{center}

\vspace{0.3em}

\noindent
\textbf{This supplementary material is organized as follows:}

\begin{itemize}
    \setlength{\itemsep}{1pt}
    \setlength{\parsep}{0pt}
    \setlength{\parskip}{0pt}
    \setlength{\topsep}{3pt}

    \item Appendix~\ref{More experiments} provides experimental
    settings, extended quantitative and qualitative comparisons,
    and analyses of frequency-extraction memory usage and
    frequency-aware reconstruction errors.

    \item Appendix~\ref{sec:supp_controlled_experiments} presents
    additional baseline comparisons, matched-Gaussian-budget
    evaluations, and an extended ablation of adaptive
    primitive compactness.

    \item Appendix~\ref{sec:method_details} details learnable
    per-Gaussian scale modulation and frequency-aware mask
    generation.

    \item Appendix~\ref{sec:limitations} discusses the limitations
    of the proposed method.

    \item Appendix~\ref{sec:impact} discusses broader social
    impacts and potential applications.

\end{itemize}

\end{tcolorbox}

\section{More Experiments}
\label{More experiments}

\begin{figure}[!t]
    \centering
    \includegraphics[width=\linewidth]{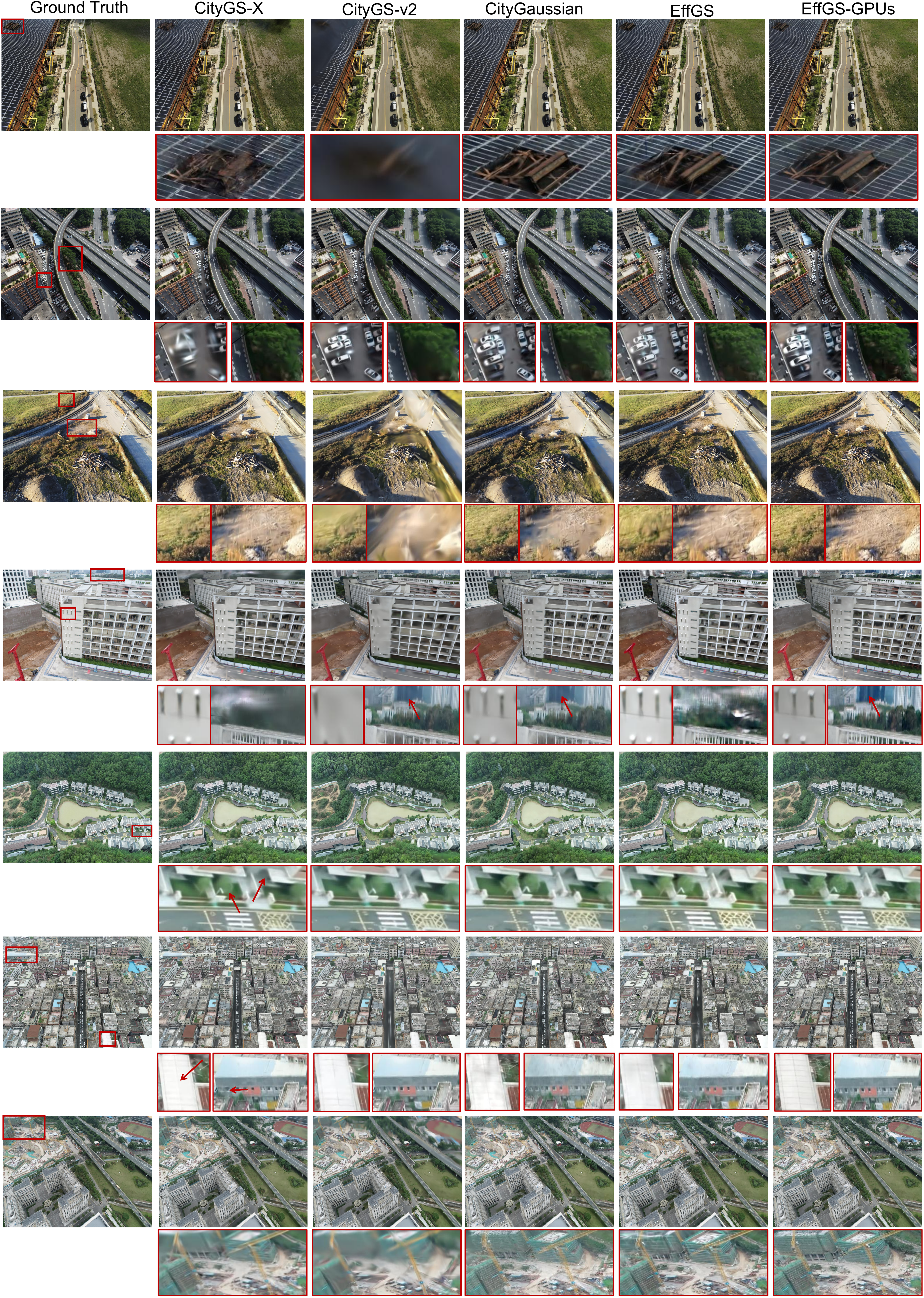} 
    \caption{Qualitative comparisons on the Mill-19, Urbanscene3D, and GauU-Scene datasets~\citep{mill,urbanscene,GauU-scene}.}
    \label{fig:render-city-sup}
    \vspace{-1em}
\end{figure}

To further demonstrate the scalability of our EffGS, we extend it to multi-GPU parallel training. We adopt the tiling strategy from CityGS~\citep{Citygs} for parallelization and conduct an extended comparison with state-of-the-art city-scale 3DGS methods~\citep{citygs-x,citygsv2,Citygs} on the Mill-19~\citep{mill}, Urbanscene3D~\citep{urbanscene}, and GauU-Scene~\citep{GauU-scene} datasets, as illustrated in Fig.~\ref{fig:render-city-sup}. We provide detailed quantitative comparisons in Tabs.~\ref{tab:ssim}, \ref{tab:psnr_comparison}, \ref{tab:lpips_comparison}, and \ref{tab:efficiency_gauu}. We also report model size and GPU memory usage, as visualized in Fig.~\ref{fig:gpu-size}. In addition, we perform two supplementary ablation studies: Memory Footprint of Frequency Extraction (Sec.~\ref{sec:memory_freq}) and Frequency Aware Error Analysis (Sec.~\ref{sec:frequency_aware_error_analysis}).

\begin{figure}[!t]
    \centering
    \includegraphics[width=\linewidth]{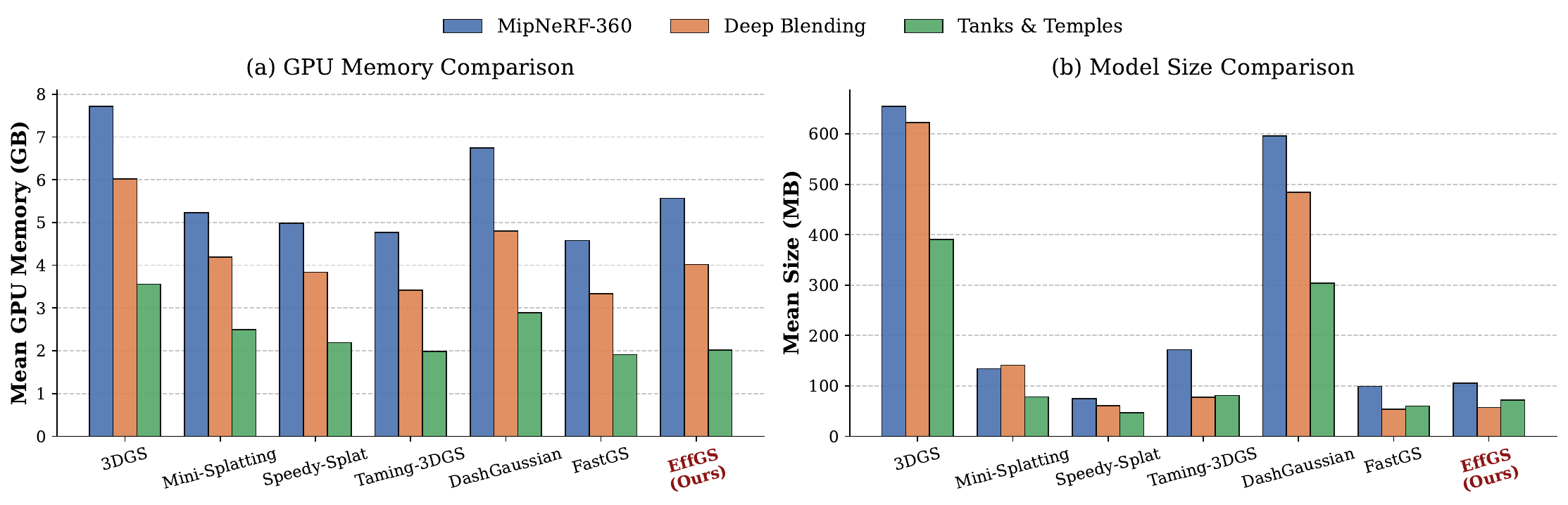} 
    \caption{Quantitative comparison of computational overhead on the Deep Blending~\citep{deep}, Mip-NeRF 360~\citep{Mip-NeRF360}, and Tanks \& Temples datasets~\citep{TNT}.}
    \label{fig:render-mip-memory-size}
    \vspace{-1em}
\end{figure}

We also provide additional qualitative (see Fig.~\ref{fig:render-mip-sup-2}) and quantitative comparisons (see Fig.~\ref{fig:render-mip-memory-size}, Tab.~\ref{tab:ssim},Tab.~\ref{tab:psnr},Tab.~\ref{tab:lpips} ) on the Deep Blending~\citep{deep}, Mip-NeRF 360~\citep{Mip-NeRF360}, and Tanks \& Temples~\citep{TNT} datasets.

\subsection{Experimental Setup and Implementation}
\label{sec:exp_details}

EffGS and the comparative baselines are implemented in
PyTorch.
Shared optimization settings, including the learning rates
for Gaussian positions and base scales, opacity reset
intervals, and density-control frequency, follow the
original 3DGS configuration.
The density-control rules and additional optimization
components of EffGS are described in
Secs.~\ref{sec:freq_importance}--\ref{Optimization}.

For bounded indoor and outdoor benchmarks, including
Mip-NeRF 360, Deep Blending, and Tanks \& Temples,
training and evaluation are conducted on a single NVIDIA
GeForce RTX 4090 with 24GB VRAM.
EffGS, Speedy-Splat~\citep{Speedy-Splat}, and
FastGS~\citep{fastgs} are evaluated under the same hardware
setting.

For large-scale urban benchmarks, we maintain the same
24GB memory constraint whenever possible.
However, several baseline configurations exceed this
capacity.
Specifically, 3DGS~\citep{3dgs},
PGSR~\citep{PGSR}, and
Mip-Splatting~\citep{mip360-splatting}
encounter out-of-memory issues, with Mip-Splatting requiring
more than 60GB of GPU memory in some scenes, and are therefore
evaluated using NVIDIA A800 GPUs.
Taming-3DGS similarly requires the A800 on several
GauU-Scene scenes.

We further distinguish the single-GPU EffGS configuration
from its distributed extension, denoted as EffGS-GPUs.
The distributed experiments use NVIDIA TITAN RTX GPUs and
adopt the tiling strategy described in
Sec.~\ref{More experiments}.
Additional memory measurements for frequency extraction
and frequency-specific error analysis are provided in
Secs.~\ref{sec:memory_freq} and
\ref{sec:frequency_aware_error_analysis}, respectively.

For the initial Gaussian primitives, we initialize
$\beta_i=0$, corresponding to a scale-modulation factor
$\gamma_i=1$.
Primitives created through cloning or splitting inherit
their parent's $\beta_i$.
The modulation parameters are optimized jointly with
the other Gaussian attributes.
The parameterization, regularization, and relation to
projected support are detailed in
Sec.~\ref{sec:compact_gaussian}.

\begin{figure}[!t]
    \centering
    \includegraphics[width=\linewidth]{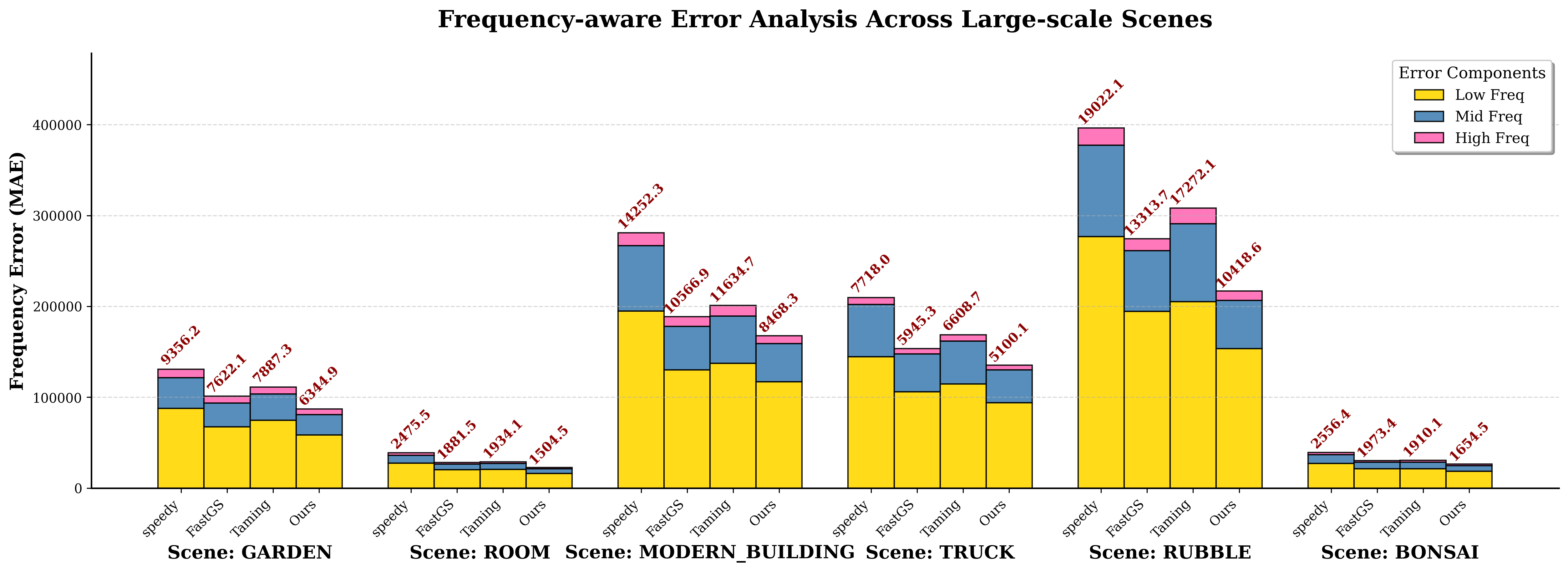}
    \caption{Low-, mid-, and high-frequency spectral magnitude
    errors across scenes.}
    \label{fig:diffreq-error}
    \vspace{-1em}
\end{figure}

\subsection{Extended Results Analysis}
\label{sec:extended_analysis}

\begin{figure*}[t]
    \centering
    \includegraphics[width=\linewidth]{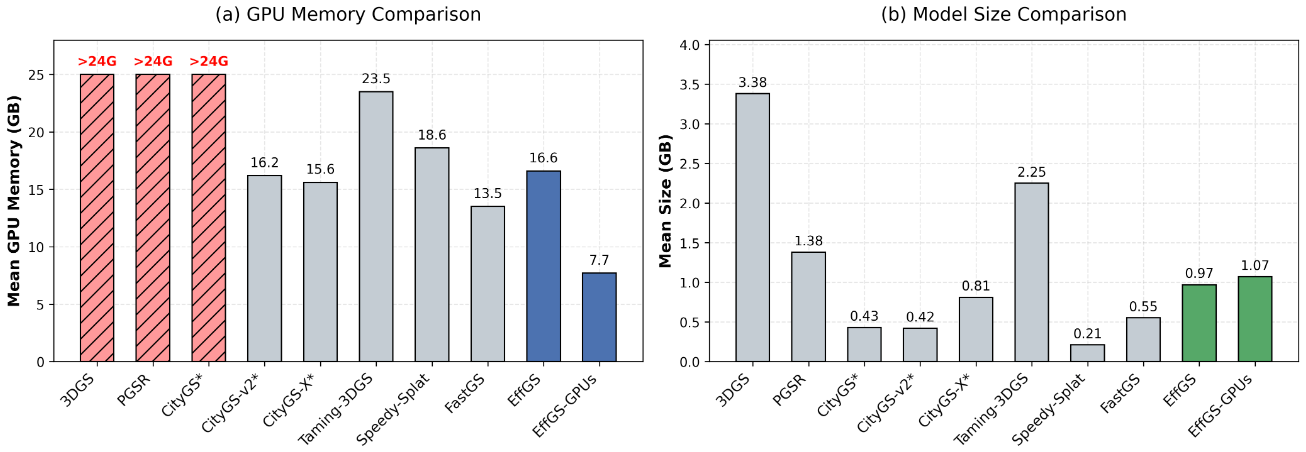} 
    \caption{Quantitative comparison of computational overhead on the Mill-19~\citep{mill}, Urbanscene3D~\citep{urbanscene}, and GauU-Scene datasets~\citep{GauU-scene}. $^\ast$ indicates multi-GPU training.}
    \label{fig:gpu-size}
    \vspace{-1em}
\end{figure*}

\paragraph{Performance on Massive Urban Scenes.}
The extended comparisons
(Tabs.~\ref{tab:ssim_comparison}--\ref{tab:lpips_comparison})
show that EffGS-GPUs achieves competitive reconstruction quality
against specialized large-scale pipelines, including CityGaussian,
CityGS-v2, and CityGS-X, with leading results across multiple
scenes and metrics.
Table~\ref{tab:efficiency_gauu} reports the training time, primitive
count, and rendering speed of EffGS and the listed baselines.
Fig.~\ref{fig:gpu-size} additionally compares memory usage and
model size, with multi-GPU configurations marked separately.

\begin{figure}[!t]
    \centering
    \includegraphics[width=\linewidth]{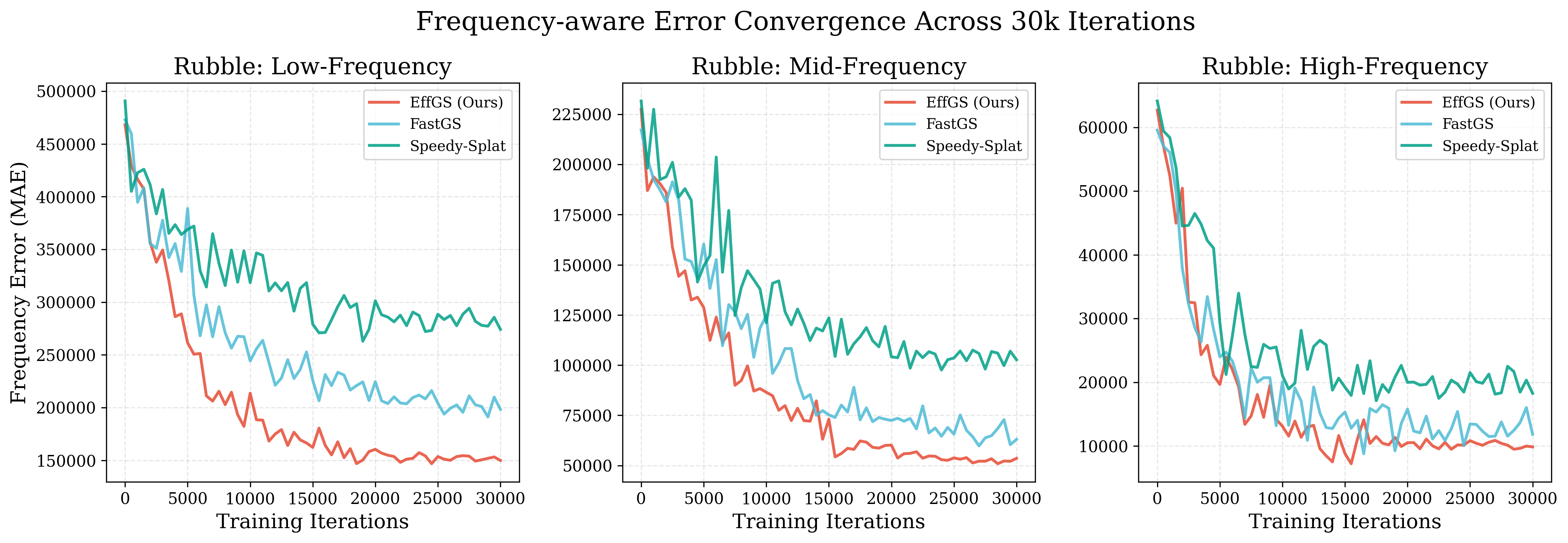}
    \caption{Band-wise spectral magnitude errors over
    30,000 training iterations on Rubble.}
    \label{fig:rubble-error}
    \vspace{-1em}
\end{figure}

Fig.~\ref{fig:rubble-error} tracks the band-wise spectral
magnitude errors on the Rubble scene.
The lower errors observed for EffGS across the three bands
complement the standard image-quality metrics and provide
additional evidence of improved spectral magnitude reconstruction.

We further evaluate LDP by incorporating it into FastGS
(Tab.~\ref{tab:fastgs+LDP}).
On the Russian scene, FastGS+LDP improves PSNR from 23.48 to
23.89 and reduces LPIPS from 0.294 to 0.257.
This comparison supports the benefit of restricting density
operations to the sampled-view active set within the FastGS
pipeline.

\begin{figure}[!t]
    \centering
    \includegraphics[width=\linewidth]{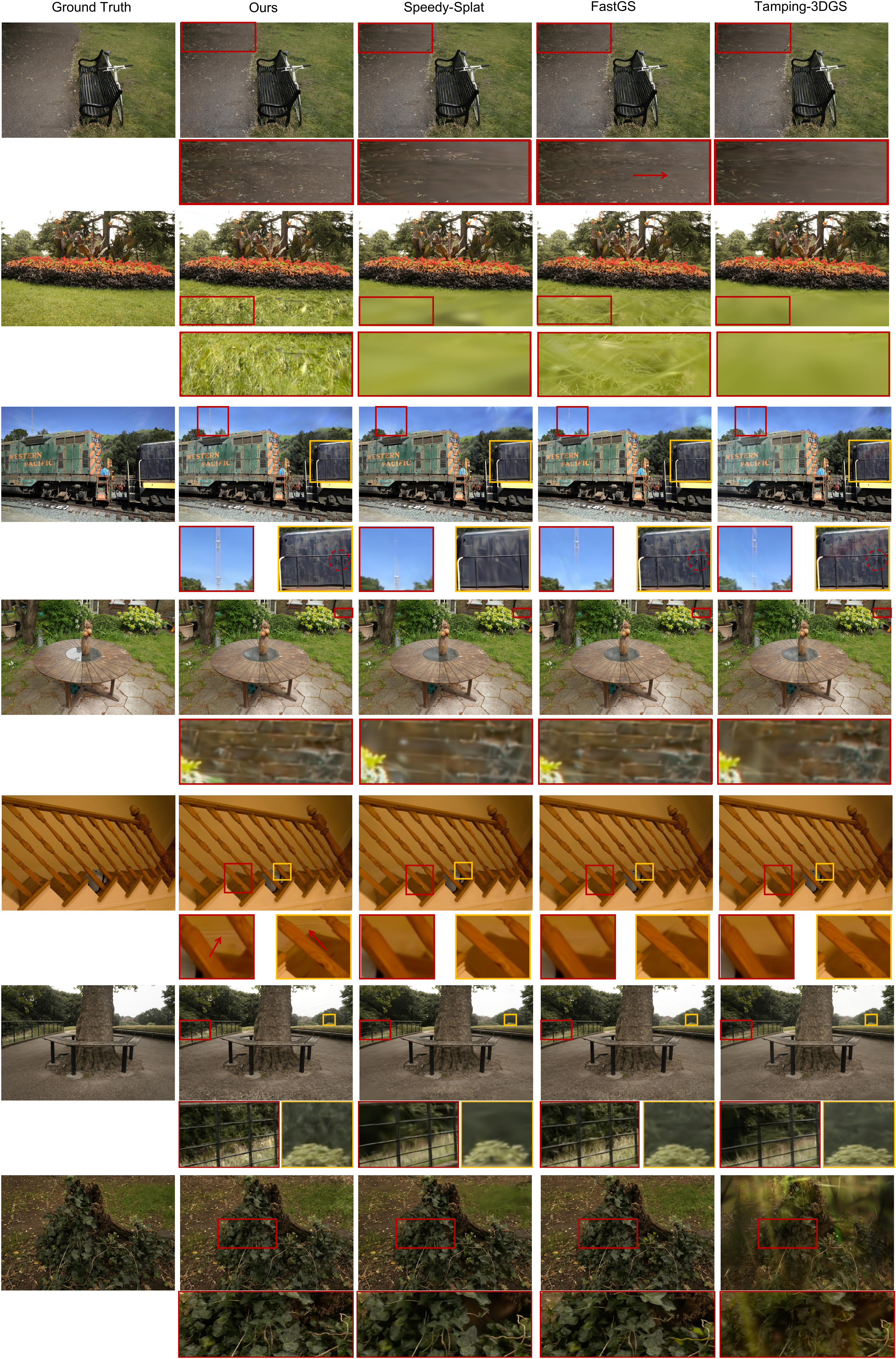} 
    \caption{Qualitative comparisons on the Deep Blending, Mip-NeRF 360, and Tanks \& Temples datasets~\citep{deep,Mip-NeRF360,TNT}.}
    \label{fig:render-mip-sup-2}
    \vspace{-1em}
\end{figure}

\paragraph{Performance on Bounded Scenes.}
The per-scene results in
Tabs.~\ref{tab:psnr}, \ref{tab:ssim}, and \ref{tab:lpips}
provide a more detailed view of reconstruction quality across
bounded scenes.
Together with the training-time comparison in
Tab.~\ref{tab:comparison}, they support the effectiveness
of EffGS across indoor and outdoor environments.
The qualitative examples in Fig.~\ref{fig:render-mip-sup-2}
show clearer edges and finer texture details than FastGS
and Speedy-Splat in the illustrated regions, complementing
the quantitative comparisons.

\subsection{Memory Footprint of Frequency Extraction}
\label{sec:memory_freq}

The frequency-aware mask is computed from two Gaussian-filtered
versions of a downsampled grayscale image.
The absolute difference is upsampled, normalized, and thresholded
according to the two-stage schedule in Sec.~\ref{sec:hf_mask}.

\begin{figure*}[t]
    \centering
    \includegraphics[width=\linewidth]{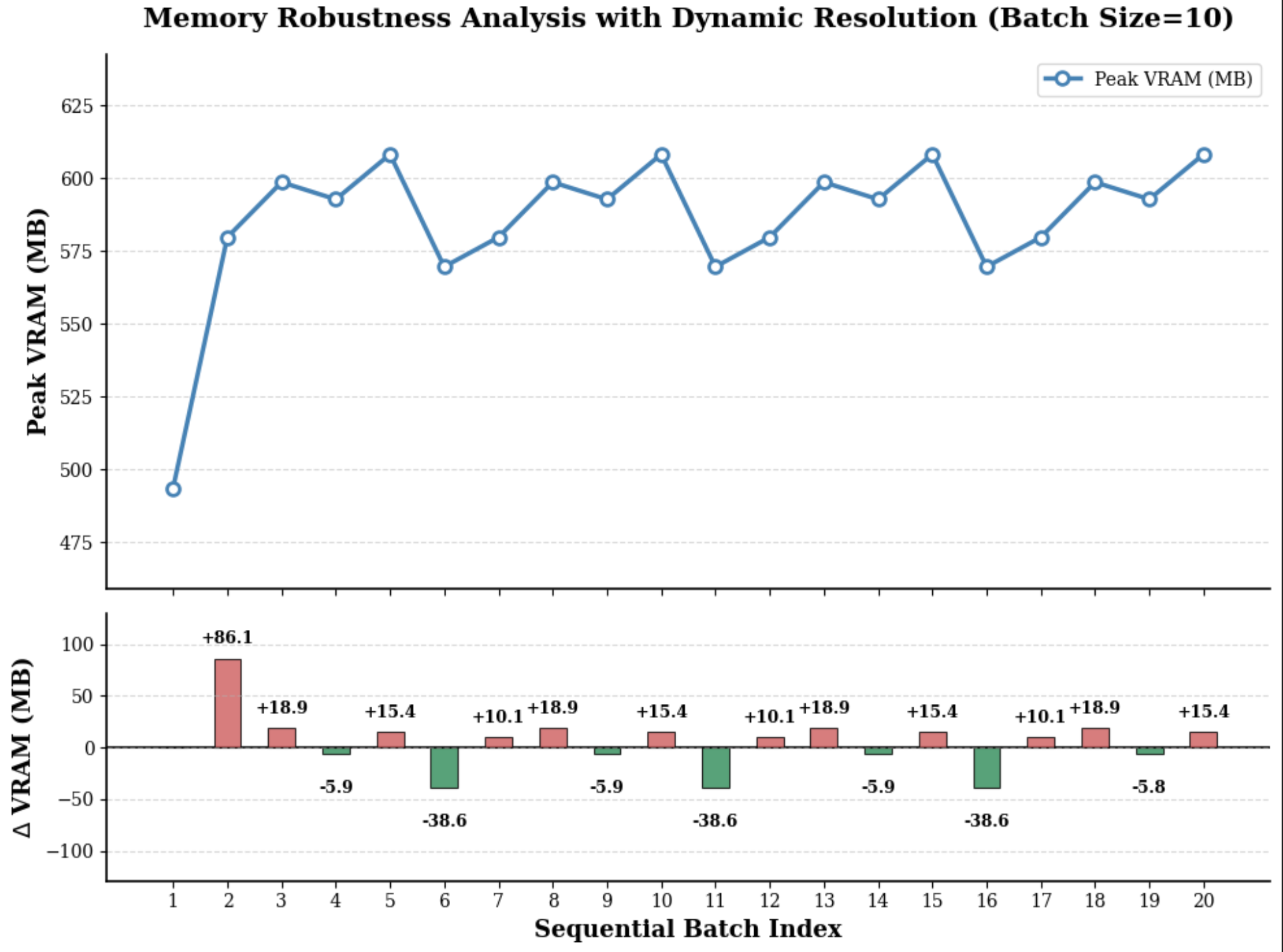}
    \caption{Peak VRAM usage (top) and step-wise changes
    (bottom) during frequency extraction with varying
    resolutions and batch size 10 on
    Russian~\citep{GauU-scene}.}
    \label{fig:freq-memory}
    \vspace{-1em}
\end{figure*}

Fig.~\ref{fig:freq-memory} evaluates memory usage during
sequential processing with varying input resolutions and a fixed
batch size of 10 on the Russian scene.
After the initial increase, peak VRAM usage fluctuates within
a bounded range without a sustained upward trend over the
tested sequence.
The step-wise changes reflect these fluctuations and support
the memory stability of frequency extraction under the
evaluated settings.

\begin{table*}[t]
\centering
\caption{
\textbf{Quantitative PSNR comparison on Mip-NeRF 360~\citep{Mip-NeRF360}.}
The best, second-best, and third-best results are indicated by
light red, light orange, and light yellow backgrounds, respectively.
}
\label{tab:psnr}

\setlength{\tabcolsep}{4pt}
\resizebox{\textwidth}{!}{
\begin{tabular}{lccccccccc}
\toprule
Method
& Bicycle
& Flowers
& Garden
& Stump
& Treehill
& Room
& Counter
& Kitchen
& Bonsai \\
\midrule

3DGS
& 25.14
& 21.30
& 27.34
& 26.64
& 22.59
& 31.71
& \second{29.16}
& 31.54
& \third{32.37} \\

Mini-Splatting
& \third{25.23}
& \third{21.43}
& 27.36
& \third{26.80}
& \third{22.76}
& 31.48
& 28.65
& 31.05
& 31.24 \\

Speedy-Splat
& 24.79
& 21.21
& 26.69
& 26.67
& 22.48
& 30.83
& 28.22
& 30.09
& 31.16 \\

Taming-3DGS
& 24.72
& 21.10
& \third{27.42}
& 26.05
& \second{22.92}
& 31.64
& \best{29.20}
& \third{31.84}
& \best{32.40} \\

DashGaussian
& \best{25.31}
& \second{21.78}
& \best{27.57}
& \second{27.17}
& \best{22.94}
& \third{31.81}
& 29.11
& 31.69
& 32.15 \\

FastGS
& 24.84
& 21.21
& 27.20
& 26.65
& \best{22.94}
& \second{31.98}
& \third{29.15}
& \second{31.87}
& 32.19 \\

Ours
& \second{25.25}
& \best{22.18}
& \second{27.45}
& \best{27.21}
& 22.34
& \best{32.03}
& \third{29.15}
& \best{31.91}
& \second{32.39} \\

\bottomrule
\end{tabular}
}
\end{table*}

\begin{table*}[t]
\centering
\caption{
\textbf{Quantitative SSIM comparison on Mip-NeRF 360~\citep{Mip-NeRF360}.}
The best, second-best, and third-best results are indicated by
light red, light orange, and light yellow backgrounds, respectively.
}
\label{tab:ssim}

\setlength{\tabcolsep}{4pt}
\resizebox{\textwidth}{!}{
\begin{tabular}{lccccccccc}
\toprule
Method
& Bicycle
& Flowers
& Garden
& Stump
& Treehill
& Room
& Counter
& Kitchen
& Bonsai \\
\midrule

3DGS
& 0.748
& 0.586
& \second{0.857}
& 0.768
& 0.636
& \third{0.927}
& \best{0.915}
& \second{0.932}
& \best{0.946} \\

Mini-Splatting
& \second{0.764}
& \best{0.614}
& 0.806
& \best{0.839}
& \best{0.656}
& \second{0.928}
& \third{0.911}
& \third{0.930}
& 0.943 \\

Speedy-Splat
& 0.704
& 0.560
& 0.814
& 0.765
& 0.590
& 0.903
& 0.876
& 0.895
& 0.925 \\

Taming-3DGS
& 0.693
& 0.552
& \third{0.851}
& 0.729
& 0.628
& 0.917
& 0.909
& 0.929
& 0.942 \\

DashGaussian
& \third{0.763}
& \third{0.604}
& \second{0.857}
& \third{0.783}
& \third{0.640}
& 0.924
& \third{0.911}
& 0.927
& \second{0.945} \\

FastGS
& 0.714
& 0.560
& 0.836
& 0.756
& 0.612
& 0.920
& 0.907
& 0.929
& 0.942 \\

Ours
& \best{0.773}
& \second{0.610}
& \best{0.861}
& \second{0.825}
& \second{0.642}
& \best{0.931}
& \second{0.913}
& \best{0.934}
& \third{0.944} \\

\bottomrule
\end{tabular}
}
\end{table*}

\begin{table*}[t]
\centering
\caption{
\textbf{Quantitative LPIPS comparison on Mip-NeRF 360~\citep{Mip-NeRF360}.}
The best, second-best, and third-best results are indicated by
light red, light orange, and light yellow backgrounds, respectively.
}
\label{tab:lpips}

\setlength{\tabcolsep}{4pt}
\resizebox{\textwidth}{!}{
\begin{tabular}{lccccccccc}
\toprule
Method
& Bicycle
& Flowers
& Garden
& Stump
& Treehill
& Room
& Counter
& Kitchen
& Bonsai \\
\midrule

3DGS
& 0.242
& \third{0.360}
& \best{0.122}
& 0.244
& 0.347
& \third{0.197}
& \third{0.183}
& \second{0.116}
& \third{0.180} \\

Mini-Splatting
& \third{0.241}
& \best{0.341}
& 0.215
& \best{0.161}
& \best{0.326}
& \second{0.190}
& \second{0.181}
& \third{0.120}
& \second{0.177} \\

Speedy-Splat
& 0.333
& 0.418
& 0.214
& 0.288
& 0.462
& 0.258
& 0.259
& 0.195
& 0.228 \\

Taming-3DGS
& 0.332
& 0.416
& \third{0.138}
& 0.324
& 0.395
& 0.227
& 0.200
& 0.128
& 0.193 \\

DashGaussian
& \best{0.222}
& \best{0.341}
& \second{0.131}
& \second{0.229}
& \second{0.333}
& 0.205
& 0.191
& 0.129
& \third{0.180} \\

FastGS
& 0.310
& 0.406
& 0.174
& 0.297
& 0.429
& 0.217
& 0.204
& 0.127
& 0.191 \\

Ours
& \second{0.240}
& \second{0.353}
& 0.162
& \third{0.243}
& \third{0.335}
& \best{0.104}
& \best{0.171}
& \best{0.106}
& \best{0.151} \\

\bottomrule
\end{tabular}
}
\end{table*}

\begin{table}[htbp]
\centering
\caption{\textbf{Quantitative results (Time) on Mip-NeRF 360~\citep{Mip-NeRF360}.}}
\label{tab:time}
\begin{tabular}{lccccccccc}
\toprule
Method & bicycle & flowers & garden & stump & treehill & room & counter & kitchen & bonsai \\
\midrule
3DGS         & 27.97 & 18.98 & 26.78 & 21.77 & 20.33 & 18.78 & 17.58 & 21.30 & 14.90 \\
Mini-Splatting & 16.17 & 17.22 & 15.97 & 16.52 & 17.05 & 18.00 & 9.83  & 10.23 & 11.02 \\
Speedy-Splat & 15.87 & 13.38 & 15.73 & 13.77 & 12.90 & 12.05 & 12.10 & 13.25 & 11.37 \\
Taming-3DGS  & 5.65  & 4.97  & 9.82  & 3.93  & 5.37  & 3.88  & 4.60  & 3.48  & 4.60  \\
DashGaussian & 9.93  & 7.05  & 8.27  & 6.57  & 8.20  & 4.00  & 3.95  & 5.52  & 3.95  \\
FastGS       & 1.92  & 1.95  & 2.47  & 1.72  & 1.72  & 1.62  & 1.83  & 2.42  & 1.83  \\
Ours         & 3.14  & 2.45  & 3.23  & 4.15  & 2.56  & 2.52  & 2.93  & 3.22  & 3.43  \\
\bottomrule
\end{tabular}
\end{table}

\begin{table*}[t]
\centering
\caption{
Quantitative SSIM comparison ($\uparrow$) across different datasets.
Building and Rubble are from Mill-19~\citep{mill};
Residence and Sci-Art are from UrbanScene3D~\citep{urbanscene};
Russian Building, Residence+, and Modern Building are from
GauU-Scene~\citep{GauU-scene}.
The best, second-best, and third-best results are indicated by
light red, light orange, and light yellow backgrounds, respectively.
$\dagger$ denotes results obtained without decoupled appearance encoding,
while * denotes results obtained without depth supervision.
}
\label{tab:ssim_comparison}

\setlength{\tabcolsep}{4pt}
\resizebox{\textwidth}{!}{
\begin{tabular}{l cc cc ccc}
\toprule
\multirow{2}{*}{Method}
& \multicolumn{2}{c}{Mill-19}
& \multicolumn{2}{c}{UrbanScene3D}
& \multicolumn{3}{c}{GauU-Scene} \\
\cmidrule(lr){2-3}
\cmidrule(lr){4-5}
\cmidrule(lr){6-8}
& Building
& Rubble
& Residence
& Sci-Art
& Russian Building
& Residence+
& Modern Building \\
\midrule

VastGaussian$\dagger$
& 0.725
& 0.745
& 0.712
& 0.765
& 0.781
& 0.738
& \third{0.789} \\

CityGaussian
& \second{0.776}
& \best{0.814}
& \second{0.810}
& \second{0.835}
& \second{0.801}
& \second{0.755}
& \second{0.791} \\

CityGS-v2
& 0.661
& 0.724
& 0.771
& 0.808
& 0.792
& 0.741
& 0.762 \\

CityGS-X*
& \third{0.771}
& \third{0.803}
& \third{0.802}
& \third{0.829}
& \third{0.793}
& 0.742
& 0.774 \\

\midrule

3DGS
& 0.722
& 0.748
& 0.782
& 0.743
& 0.770
& 0.686
& 0.751 \\

PGSR
& 0.482
& 0.723
& 0.758
& 0.802
& 0.759
& 0.315
& 0.705 \\

Mip-Splatting
& 0.683
& 0.722
& 0.749
& 0.806
& 0.751
& 0.518
& 0.710 \\

\midrule

Taming-3DGS
& 0.498
& 0.603
& 0.560
& 0.706
& 0.738
& 0.661
& 0.744 \\

Speedy-Splat
& 0.462
& 0.581
& 0.649
& 0.667
& 0.697
& 0.569
& 0.671 \\

FastGS
& 0.654
& 0.669
& 0.688
& 0.728
& 0.741
& 0.659
& 0.722 \\

\midrule

EffGS
& 0.728
& 0.783
& 0.777
& 0.778
& 0.768
& \third{0.747}
& 0.754 \\

EffGS-GPUs
& \best{0.782}
& \second{0.811}
& \best{0.813}
& \best{0.844}
& \best{0.803}
& \best{0.786}
& \best{0.795} \\

\bottomrule
\end{tabular}
}
\end{table*}

\begin{table*}[t]
\centering
\caption{
Quantitative PSNR comparison ($\uparrow$) across different datasets.
Building and Rubble are from Mill-19~\citep{mill};
Residence and Sci-Art are from UrbanScene3D~\citep{urbanscene};
Russian Building, Residence+, and Modern Building are from
GauU-Scene~\citep{GauU-scene}.
The best, second-best, and third-best results are indicated by
light red, light orange, and light yellow backgrounds, respectively.
$\dagger$ denotes results obtained without decoupled appearance encoding,
while * denotes results obtained without depth supervision.
}
\label{tab:psnr_comparison}

\setlength{\tabcolsep}{4pt}
\resizebox{\textwidth}{!}{
\begin{tabular}{l cc cc ccc}
\toprule
\multirow{2}{*}{Method}
& \multicolumn{2}{c}{Mill-19}
& \multicolumn{2}{c}{UrbanScene3D}
& \multicolumn{3}{c}{GauU-Scene} \\
\cmidrule(lr){2-3}
\cmidrule(lr){4-5}
\cmidrule(lr){6-8}
& Building
& Rubble
& Residence
& Sci-Art
& Russian Building
& Residence+
& Modern Building \\
\midrule

VastGaussian$\dagger$
& \third{21.83}
& 25.24
& 21.06
& \second{22.59}
& 23.98
& 23.41
& 25.53 \\

CityGaussian
& 21.56
& \second{25.79}
& \second{22.03}
& \third{22.42}
& \third{24.11}
& \third{23.65}
& \second{26.01} \\

CityGS-v2
& 19.85
& 24.02
& 21.23
& 20.71
& 24.04
& 23.43
& 25.78 \\

CityGS-X*
& 21.78
& 25.45
& \best{22.11}
& 22.32
& \second{24.18}
& \best{23.91}
& \third{25.95} \\

\midrule

3DGS
& 20.62
& \third{25.49}
& 21.45
& 21.92
& 23.74
& 22.10
& 25.15 \\

PGSR
& 17.01
& 23.23
& 20.58
& 19.43
& 23.30
& 14.65
& 23.77 \\

Mip-Splatting
& 20.65
& 24.05
& 20.98
& 21.75
& 22.48
& 17.97
& 23.76 \\

\midrule

Taming-3DGS
& 19.13
& 23.31
& 19.01
& 20.47
& 23.51
& 22.34
& 25.46 \\

Speedy-Splat
& 16.90
& 22.50
& 20.04
& 18.99
& 22.87
& 20.32
& 24.08 \\

FastGS
& 20.88
& 23.76
& 20.13
& 20.34
& 23.48
& 22.12
& 24.81 \\

\midrule

EffGS
& \second{22.05}
& 25.31
& 21.23
& 21.77
& 24.05
& 23.28
& 25.35 \\

EffGS-GPUs
& \best{22.34}
& \best{25.83}
& \third{21.89}
& \best{22.73}
& \best{24.34}
& \second{23.89}
& \best{26.14} \\

\bottomrule
\end{tabular}
}
\end{table*}

\begin{figure*}[t]
    \centering
    \includegraphics[width=\textwidth]{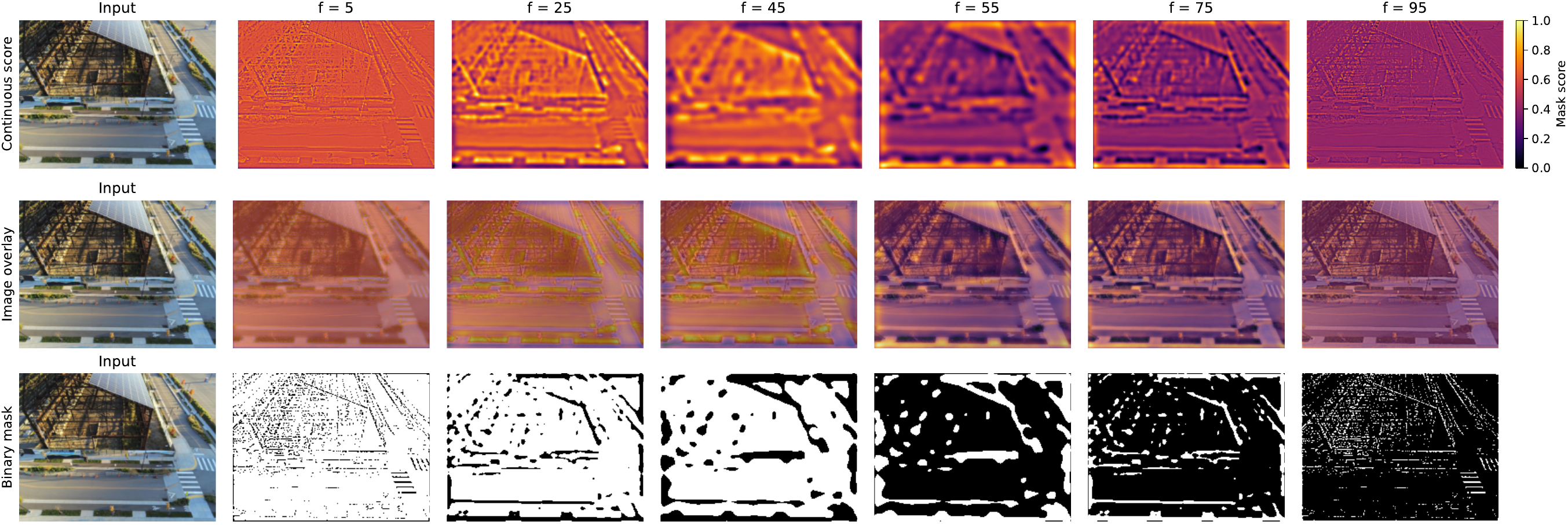}
    \caption{
    Visualization of the frequency-aware mask at different schedule
    parameters $f$ on a fixed input image.
    Top: continuous mask scores before thresholding.
    Middle: score heatmaps overlaid on the input.
    Bottom: binary masks obtained with a threshold of 0.5,
    where white indicates selected pixels.
    Scores are normalized independently for each setting.
    }
    \label{fig:frequency_aware_mask}
\end{figure*}

\begin{table*}[t]
\centering
\caption{
Quantitative LPIPS comparison ($\downarrow$) across different datasets.
Building and Rubble are from Mill-19~\citep{mill};
Residence and Sci-Art are from UrbanScene3D~\citep{urbanscene};
Russian Building, Residence+, and Modern Building are from
GauU-Scene~\citep{GauU-scene}.
The best, second-best, and third-best results are indicated by
light red, light orange, and light yellow backgrounds, respectively.
$\dagger$ denotes results obtained without decoupled appearance encoding,
while * denotes results obtained without depth supervision.
}
\label{tab:lpips_comparison}

\setlength{\tabcolsep}{4pt}
\resizebox{\textwidth}{!}{
\begin{tabular}{l cc cc ccc}
\toprule
\multirow{2}{*}{Method}
& \multicolumn{2}{c}{Mill-19}
& \multicolumn{2}{c}{UrbanScene3D}
& \multicolumn{3}{c}{GauU-Scene} \\
\cmidrule(lr){2-3}
\cmidrule(lr){4-5}
\cmidrule(lr){6-8}
& Building
& Rubble
& Residence
& Sci-Art
& Russian Building
& Residence+
& Modern Building \\
\midrule

VastGaussian$\dagger$
& 0.271
& 0.268
& 0.263
& 0.259
& 0.251
& 0.301
& \second{0.245} \\

CityGaussian
& \third{0.267}
& \third{0.226}
& \second{0.213}
& \second{0.232}
& \second{0.223}
& \third{0.289}
& 0.253 \\

CityGS-v2
& 0.382
& 0.311
& 0.235
& 0.265
& \third{0.234}
& 0.294
& 0.254 \\

CityGS-X*
& \second{0.256}
& \second{0.224}
& \third{0.221}
& \third{0.241}
& \best{0.220}
& \second{0.284}
& \best{0.234} \\

\midrule

3DGS
& 0.304
& 0.279
& 0.238
& 0.265
& \third{0.234}
& 0.321
& \third{0.249} \\

PGSR
& 0.541
& 0.352
& 0.292
& 0.277
& 0.278
& 0.657
& 0.327 \\

Mip-Splatting
& 0.344
& 0.324
& 0.272
& 0.259
& 0.252
& 0.482
& 0.290 \\

\midrule

Taming-3DGS
& 0.542
& 0.477
& 0.509
& 0.410
& 0.291
& 0.352
& 0.282 \\

Speedy-Splat
& 0.593
& 0.501
& 0.411
& 0.452
& 0.352
& 0.469
& 0.382 \\

FastGS
& 0.380
& 0.402
& 0.351
& 0.351
& 0.294
& 0.377
& 0.315 \\

\midrule

EffGS
& 0.276
& 0.282
& 0.265
& 0.297
& 0.254
& 0.291
& 0.273 \\

EffGS-GPUs
& \best{0.253}
& \best{0.217}
& \best{0.208}
& \best{0.224}
& 0.236
& \best{0.281}
& 0.256 \\

\bottomrule
\end{tabular}
}
\end{table*}

\begin{table}[h]
\centering
\caption{Ablation study on the Russian scene from the GauU-Scene dataset~\citep{GauU-scene}.}
\label{tab:fastgs+LDP}
\begin{tabular}{l ccc}
\toprule
Method & SSIM $\uparrow$ & PSNR $\uparrow$ & LPIPS $\downarrow$ \\
\midrule
FastGS     & 0.741 & 23.48 & 0.294 \\
FastGS+LDP & 0.758 & 23.89 &  0.257 \\
\bottomrule
\end{tabular}
\end{table}

\begin{table*}[t]
\centering
\caption{Efficiency comparison on Residence+, Russian Building, and Modern Building scenes from the GauU-Scene dataset~\citep{GauU-scene}.}
\label{tab:efficiency_gauu}
\resizebox{\textwidth}{!}{
\begin{tabular}{l ccc ccc ccc}
\toprule
\multirow{2}{*}{Method} & \multicolumn{3}{c}{Residence+} & \multicolumn{3}{c}{Russian Building} & \multicolumn{3}{c}{Modern Building} \\
\cmidrule(lr){2-4} \cmidrule(lr){5-7} \cmidrule(lr){8-10}
& Time (min) & GS (M) & FPS & Time (min) & GS (M) & FPS & Time (min) & GS (M) & FPS \\
\midrule
CityGaussian & 261 & 8.14 & 68  & 214 & 7.02 & 55  & 217 & 7.90 & 57 \\
CityGS-v2 & 203 & 8.04 & 46  & 182 & 6.97 & 33  & 188 & 7.90 & 35 \\
Speedy-splat  & 50  & 0.50 & 195 & 43  & 0.82 & 189 & 43  & 0.87 & 170 \\
FastGS        & 14  & 2.91 & 252 & 12  & 1.89 & 231 & 12  & 2.23 & 224 \\
EffGS         & 19  & 4.86 & 131 & 21  & 3.45 & 127 & 23  & 4.37 & 115 \\
\bottomrule
\end{tabular}
}
\end{table*}

\subsection{Frequency-Aware Error Analysis}
\label{sec:frequency_aware_error_analysis}

We visualize the spatial masks used for training guidance and
measure spectral magnitude discrepancies in separate frequency
bands.
These analyses complement PSNR, SSIM, LPIPS, and qualitative
renderings by characterizing mask selection and frequency-specific
differences.

\subsubsection{Frequency-Aware Mask Visualization}

Fig.~\ref{fig:frequency_aware_mask} shows continuous selection
scores, their overlays, and binary masks on a fixed image at
different schedule parameters $f$.
For $f\geq50$, increasing $f$ reduces the Gaussian filter scales,
producing responses around progressively finer structures.
For $f<50$, the normalized response is complemented before
thresholding.
The heatmaps represent spatial selection scores, and the
binary masks identify pixels selected at the threshold of 0.5.
Because scores are normalized independently for each setting,
their colors indicate relative selection strength within
each image.

\subsubsection{Spectral Decomposition and Error Formulation}
\label{sec:spectral_analysis}
We compute the centered Fourier spectra of the grayscale
ground-truth and rendered images. For
$I\in\{I_{\mathrm{gt}}, I_{\mathrm{render}}\}$,
\begin{equation}
\widehat{I}
=
\operatorname{fftshift}\!\left(\mathcal{F}(I)\right),
\qquad
A_I = |\widehat{I}|,
\end{equation}
where $\mathcal{F}$ denotes the 2D discrete Fourier transform,
$\widehat{I}$ is the centered complex spectrum, and $A_I$ is
its magnitude spectrum.

Let $(\xi,\eta)$ denote a frequency-grid coordinate and
$D(\xi,\eta)$ its Euclidean radial distance from the centered
zero-frequency location. We partition the frequency coordinates
into three bands:
\begin{equation}
\begin{aligned}
\Omega_{\mathrm{low}}
&= \{(\xi,\eta): D(\xi,\eta) \leq 30\},\\
\Omega_{\mathrm{mid}}
&= \{(\xi,\eta): 30 < D(\xi,\eta) \leq 80\},\\
\Omega_{\mathrm{high}}
&= \{(\xi,\eta): D(\xi,\eta) > 80\}.
\end{aligned}
\end{equation}
The thresholds are measured in frequency-grid pixels and
therefore define bands relative to the image resolution used
in each evaluation. For each band
$b\in\{\mathrm{low},\mathrm{mid},\mathrm{high}\}$, we compute
the mean absolute spectral magnitude error:
\begin{equation}
E_b =
\frac{1}{|\Omega_b|}
\sum_{(\xi,\eta)\in\Omega_b}
\left|
A_{I_{\mathrm{render}}}(\xi,\eta)
-
A_{I_{\mathrm{gt}}}(\xi,\eta)
\right|.
\label{eq:freq_band_error}
\end{equation}
This metric measures differences in Fourier magnitude.
Since it does not capture phase differences, we interpret it
together with the standard image-quality metrics and
qualitative results.

\subsubsection{Spatial Visualization and Quantitative Comparison}

To visualize high-frequency response discrepancies, we define
the binary spectral mask
\begin{equation}
M_{\mathrm{high}}(\xi,\eta)
=
\mathbb{I}\!\left[(\xi,\eta)\in\Omega_{\mathrm{high}}\right],
\end{equation}
where $\mathbb{I}[\cdot]$ denotes the indicator function.
We apply this mask to the complex spectrum and compute the
magnitude of the reconstructed high-pass response:
\begin{equation}
S_{\mathrm{hf}}(I)
=
\left|
\mathcal{F}^{-1}\!\left(
\operatorname{ifftshift}\!\left(
\widehat{I}\odot M_{\mathrm{high}}
\right)
\right)
\right|,
\label{eq:spatial_high_freq}
\end{equation}
where $\mathcal{F}^{-1}$ denotes the inverse 2D discrete
Fourier transform and $\odot$ denotes element-wise
multiplication.

We visualize
\begin{equation}
\Delta S_{\mathrm{hf}}
=
\left|
S_{\mathrm{hf}}(I_{\mathrm{gt}})
-
S_{\mathrm{hf}}(I_{\mathrm{render}})
\right|
\end{equation}
after normalization by its 99th percentile. The resulting map
shows spatial differences in high-pass response magnitude,
rather than squared-energy error. Percentile normalization
emphasizes the spatial distribution of these discrepancies.

We also present the three band-wise errors as stacked bars,
allowing comparisons between methods within each frequency
band. The stacked height is the sum of the three band-wise
mean errors; it is neither the overall pixel-space error nor
a decomposition of total error into frequency-band proportions.

\section{Additional Comparisons and Controlled Experiments}
\label{sec:supp_controlled_experiments}

We compare EffGS with EDGS and 3DGS-MCMC under reported and
matched-budget settings, and further evaluate adaptive
primitive compactness.

\subsection{Reference and Matched-Budget Comparisons}
\label{sec:supp_reference_and_matched}

Tab.~\ref{tab:supp_reference_and_matched} compares reconstruction
quality and training time on Deep Blending and Tanks \& Temples.
Under reported settings, EffGS has lower reported training times
on both datasets and achieves the highest SSIM and PSNR on Deep
Blending, while EDGS achieves the lowest LPIPS on both datasets
and the highest SSIM and PSNR on Tanks \& Temples.
Under Gaussian budgets matched to EffGS---0.25M on Deep Blending
and 0.44M on Tanks \& Temples---EffGS achieves the best PSNR,
SSIM, LPIPS, and training time on both datasets.
These results show that its quality advantages persist under
matched budgets rather than relying on more Gaussian primitives.

\begin{table*}[t]
\centering
\caption{
Comparison on Deep Blending and Tanks \& Temples.
Top: reported settings.
Bottom: Gaussian budgets matched to EffGS.
Time is in minutes and $N_{\mathrm{GS}}$ in millions;
-- denotes unavailable counts.
Best quality and timing results within each block and dataset are bold.
}
\label{tab:supp_reference_and_matched}

\setlength{\tabcolsep}{3pt}
\resizebox{\textwidth}{!}{%
\begin{tabular}{l ccccc ccccc}
\toprule
\multirow{2}{*}{Method}
& \multicolumn{5}{c}{Deep Blending}
& \multicolumn{5}{c}{Tanks \& Temples} \\
\cmidrule(lr){2-6}
\cmidrule(lr){7-11}
& Time $\downarrow$
& PSNR $\uparrow$
& SSIM $\uparrow$
& LPIPS $\downarrow$
& $N_{\mathrm{GS}}$
& Time $\downarrow$
& PSNR $\uparrow$
& SSIM $\uparrow$
& LPIPS $\downarrow$
& $N_{\mathrm{GS}}$ \\
\midrule

\multicolumn{11}{l}{
\textit{Uncontrolled Gaussian‑point count}
} \\
\addlinespace[2pt]

EDGS
& 30.00
& 29.81
& 0.904
& \textbf{0.223}
& --
& 23.00
& \textbf{24.28}
& \textbf{0.868}
& \textbf{0.132}
& -- \\

3DGS-MCMC
& 19.00
& 29.56
& 0.902
& 0.244
& --
& 13.00
& 24.22
& 0.863
& 0.158
& -- \\

EffGS (Ours)
& \textbf{2.94}
& \textbf{30.01}
& \textbf{0.910}
& 0.238
& 0.25
& \textbf{1.94}
& 24.22
& 0.861
& 0.166
& 0.44 \\

\midrule

\multicolumn{11}{l}{
\textit{Matched Gaussian budgets: controlled comparison}
} \\
\addlinespace[2pt]

EDGS
& 7.00
& 28.83
& 0.877
& 0.309
& 0.25
& 6.00
& 22.19
& 0.841
& 0.194
& 0.44 \\

3DGS-MCMC
& 5.00
& 28.67
& 0.874
& 0.294
& 0.25
& 4.00
& 22.24
& 0.837
& 0.207
& 0.44 \\

EffGS (Ours)
& \textbf{2.94}
& \textbf{30.01}
& \textbf{0.910}
& \textbf{0.238}
& 0.25
& \textbf{1.94}
& \textbf{24.22}
& \textbf{0.861}
& \textbf{0.166}
& 0.44 \\

\bottomrule
\end{tabular}%
}
\end{table*}

\subsection{Matched-Budget Comparisons on Large-Scale Scenes}
\label{sec:supp_matched_budgets_large_scale}

We further evaluate matched-budget performance on Mill-19 and
UrbanScene3D.
For Taming-3DGS, we set its target Gaussian budget to match EffGS.
For the EDGS variant, denoted as EDGS+3DGS, we disable
densification and initialize optimization with the same
Gaussian budget.
These budget-constrained configurations are evaluated separately
from the methods' original operating settings.

Tab.~\ref{tab:supp_iso_budget_large_scale} summarizes the results.
At the reported matched budgets of 3.76M Gaussians on Mill-19
and 3.05M Gaussians on UrbanScene3D, EffGS achieves the best
reconstruction metrics and the shortest training time among
the evaluated methods.
On Mill-19, EffGS achieves 23.68 dB PSNR and 0.279 LPIPS,
compared with 22.57 dB and 0.372 for EDGS+3DGS.
On UrbanScene3D, EffGS achieves 21.50 dB PSNR and 0.281 LPIPS,
compared with 21.17 dB and 0.330 for EDGS+3DGS.
Together with the results on Deep Blending and Tanks \& Temples,
these comparisons support the effectiveness of EffGS under
matched primitive budgets across different scene scales.
They do not, by themselves, isolate the contribution of an
individual component of the full framework.

\begin{table*}[t]
\centering
\caption{
Matched-budget comparison on Mill-19 and UrbanScene3D.
EDGS+3DGS uses matched initialization with densification disabled.
Time is in minutes and $N_{\mathrm{GS}}$ in millions.
Best quality and timing results are bold.
}
\label{tab:supp_iso_budget_large_scale}

\setlength{\tabcolsep}{3pt}
\resizebox{\textwidth}{!}{%
\begin{tabular}{l ccccc ccccc}
\toprule
\multirow{2}{*}{Method}
& \multicolumn{5}{c}{Mill-19}
& \multicolumn{5}{c}{UrbanScene3D} \\
\cmidrule(lr){2-6}
\cmidrule(lr){7-11}
& Time $\downarrow$
& PSNR $\uparrow$
& SSIM $\uparrow$
& LPIPS $\downarrow$
& $N_{\mathrm{GS}}$
& Time $\downarrow$
& PSNR $\uparrow$
& SSIM $\uparrow$
& LPIPS $\downarrow$
& $N_{\mathrm{GS}}$ \\
\midrule

Taming-3DGS
& 57
& 22.52
& 0.693
& 0.384
& 3.76
& 61
& 21.14
& 0.719
& 0.332
& 3.05 \\

EDGS+3DGS
& 83
& 22.57
& 0.716
& 0.372
& 3.76
& 91
& 21.17
& 0.716
& 0.330
& 3.05 \\

EffGS (Ours)
& \textbf{20}
& \textbf{23.68}
& \textbf{0.755}
& \textbf{0.279}
& 3.76
& \textbf{22}
& \textbf{21.50}
& \textbf{0.778}
& \textbf{0.281}
& 3.05 \\

\bottomrule
\end{tabular}%
}
\end{table*}

\subsection{Extended Analysis of Adaptive Primitive Compactness}
\label{sec:supp_extended_compactness}

The learnable factor $\gamma_i$ provides a dedicated optimization
pathway for adaptive primitive compactness.
Although a multiplicative scale modulation does not expand
the static representational space of a Gaussian, it changes
the optimization parameterization and participates in
compactness-aware primitive control.

We evaluate its contribution on five representative scenes:
Garden, Room, and Bonsai from Mip-NeRF 360, Rubble from Mill-19,
and Modern Building from GauU-Scene.
Tab.~\ref{tab:supp_compactness_extended} reports the aggregated
results across these scenes.
Compared with the variant without learnable compactness,
the full model reduces the average errors in all three
frequency bands, improves PSNR from 26.99 to 28.51 dB and
SSIM from 0.810 to 0.855, and reduces LPIPS from 0.264 to 0.194.
The average Gaussian count also decreases from 2.30M to 2.02M.
Thus, the observed improvement in reconstruction quality is
accompanied by a more compact representation, rather than
an increase in primitive count.

\begin{table*}[t]
\centering
\caption{
Extended ablation of learnable primitive compactness,
averaged over Garden, Room, Bonsai, Rubble, and Modern Building.
$N_{\mathrm{GS}}$ is reported in millions.
The best results are shown in bold.
}
\label{tab:supp_compactness_extended}

\setlength{\tabcolsep}{4pt}
\resizebox{\textwidth}{!}{%
\begin{tabular}{l ccc ccc c}
\toprule
\multirow{2}{*}{Method}
& \multicolumn{3}{c}{Frequency Error}
& \multicolumn{3}{c}{Reconstruction Quality}
& \multirow{2}{*}{$N_{\mathrm{GS}}$ $\downarrow$} \\
\cmidrule(lr){2-4}
\cmidrule(lr){5-7}
& Low $\downarrow$
& Mid $\downarrow$
& High $\downarrow$
& PSNR $\uparrow$
& SSIM $\uparrow$
& LPIPS $\downarrow$
& \\
\midrule

w/o Learnable Compactness ($\gamma_i$)
& 86726.7
& 30836.8
& 7073.6
& 26.99
& 0.810
& 0.264
& 2.30 \\

EffGS (Ours)
& \textbf{72713.2}
& \textbf{25773.5}
& \textbf{5678.1}
& \textbf{28.51}
& \textbf{0.855}
& \textbf{0.194}
& \textbf{2.02} \\

\bottomrule
\end{tabular}%
}
\end{table*}

\paragraph{Summary.}
The matched-budget comparisons show that EffGS achieves
better reconstruction quality than the evaluated baselines
at the tested Gaussian budgets, while also reporting shorter
training times.
The extended compactness ablation further shows that the
full model improves average reconstruction quality and
frequency-specific errors while using fewer primitives.
These results provide complementary evidence for the
quality--efficiency benefits of the proposed framework.

\section{Methodological Details}
\label{sec:method_details}

\subsection{Learnable Scale Modulation for Primitive Compactness}
\label{sec:compact_gaussian}

Building upon the Compact Box mechanism in
FastGS~\citep{fastgs}, we introduce a
\textbf{learnable per-Gaussian scale factor} $\gamma_i$.
This factor reparameterizes the effective Gaussian scales,
providing an additional optimization pathway for adapting
primitive extent while retaining the existing tile-culling rule.

\paragraph{Covariance modulation.}
In vanilla 3DGS~\citep{3dgs}, the covariance of Gaussian $i$
is constructed from its base-scale matrix
$S_i=\operatorname{diag}(s_i^x,s_i^y,s_i^z)$,
with $s_i^x,s_i^y,s_i^z>0$, and rotation matrix $R_i$:
\begin{equation}
\Sigma_i
=
R_iS_iS_i^\top R_i^\top.
\label{eq:cov_3d}
\end{equation}
We augment each Gaussian with a learnable scalar
$\beta_i\in\mathbb{R}$, initialized to zero for the initial
Gaussian primitives. Its corresponding positive scale
multiplier is
\begin{equation}
\gamma_i
=
2\,\operatorname{sigmoid}(\beta_i).
\label{eq:var_approx}
\end{equation}
For finite $\beta_i$, we have $\gamma_i\in(0,2)$, with
$\gamma_i=1$ at initialization.

The effective scale matrix is
\begin{equation}
\widetilde{S}_i
=
\operatorname{diag}
(\gamma_i s_i^x,\gamma_i s_i^y,\gamma_i s_i^z)
=
\gamma_i S_i.
\label{eq:scaling_mod}
\end{equation}
Consequently,
\begin{equation}
\widetilde{\Sigma}_i
=
R_i\widetilde{S}_i\widetilde{S}_i^\top R_i^\top
=
\gamma_i^2\Sigma_i.
\end{equation}
The scalar factor changes the overall extent, while anisotropic
axis ratios remain governed by the base scales.

Let $\widetilde{\Sigma}_i^{\mathrm{2D}}$ denote the projected
covariance and $\mathbf{p}_i^{\mathrm{2D}}$ the projected
center, following the notation in the main paper.
For an image-plane coordinate $\mathbf{u}\in\mathbb{R}^2$,
the squared Mahalanobis distance is
\begin{equation}
m_i(\mathbf{u})
=
(\mathbf{u}-\mathbf{p}_i^{\mathrm{2D}})^\top
(\widetilde{\Sigma}_i^{\mathrm{2D}})^{-1}
(\mathbf{u}-\mathbf{p}_i^{\mathrm{2D}}).
\label{eq:maha_dist}
\end{equation}
Compact Box uses this quantity to determine the spatial support
relevant to tile culling.

\paragraph{Gradient flow.}
We optimize $\beta_i$ jointly with the other Gaussian
attributes using Adam. Its gradient follows the chain rule:
\begin{equation}
\frac{\partial\mathcal{L}}{\partial\beta_i}
=
\frac{\partial\mathcal{L}}{\partial\gamma_i}
\frac{\partial\gamma_i}{\partial\beta_i}.
\label{eq:grad_chain}
\end{equation}
From Eq.~\eqref{eq:var_approx},
\begin{equation}
\frac{\partial\gamma_i}{\partial\beta_i}
=
2\,\operatorname{sigmoid}(\beta_i)
\left(1-\operatorname{sigmoid}(\beta_i)\right).
\label{eq:grad_act}
\end{equation}
This differentiable parameterization allows the training
objective to update the modulation factor through the effective
Gaussian scales.

\paragraph{Impact on projected support.}
Holding the Gaussian center and camera fixed, let
$\Pi_i=J_iW$, where $W$ is the rotational part of the
world-to-camera transformation and $J_i$ is the projection
Jacobian at the center of Gaussian $i$.
Under the first-order covariance projection,
\begin{equation}
\widetilde{\Sigma}_i^{\mathrm{2D}}
=
\Pi_i\widetilde{\Sigma}_i\Pi_i^\top
=
\gamma_i^2\Pi_i\Sigma_i\Pi_i^\top
=
\gamma_i^2\Sigma_i^{\mathrm{2D}}.
\end{equation}
For a fixed squared Mahalanobis level $q>0$, let
$\ell_{i,\kappa}$ denote the $\kappa$-th eigenvalue of
$\Sigma_i^{\mathrm{2D}}$, where
$\kappa\in\{1,2\}$ indexes the two principal axes.
The corresponding ellipse semi-axis length is
\begin{equation}
\rho_{i,\kappa}
=
\sqrt{q\gamma_i^2\ell_{i,\kappa}}
=
\gamma_i\sqrt{q\ell_{i,\kappa}}.
\end{equation}
Thus, with the base covariance held fixed, a smaller factor
contracts the ideal projected support and can reduce the
number of intersected tiles.

If a fixed screen-space covariance
$\varepsilon_{\mathrm{cov}}\mathbf{I}_2$ is added during
rasterization, where $\varepsilon_{\mathrm{cov}}>0$ and
$\mathbf{I}_2$ denotes the $2\times2$ identity matrix, the
projected eigenvalues become
$\gamma_i^2\ell_{i,\kappa}+\varepsilon_{\mathrm{cov}}$.
The corresponding semi-axis lengths are
\begin{equation}
\sqrt{
q\left(
\gamma_i^2\ell_{i,\kappa}
+
\varepsilon_{\mathrm{cov}}
\right)
}.
\end{equation}
Discrete tile assignment further makes the realized tile count
change in steps as the support varies.

During training, both the base scales and $\gamma_i$ are
optimized. The resulting effective support therefore depends
on their product, while total rasterization cost also depends
on the number and arrangement of primitives.

\paragraph{Training and regularization.}
For the initial Gaussian primitives, we set $\beta_i=0$,
giving $\gamma_i=1$. Primitives created by cloning or splitting
inherit their parent's $\beta_i$, as specified in
Sec.~\ref{sec:local_densification_pruning}.
The modulation parameters are subsequently optimized with
a dedicated learning rate.
The L2 term $\mathcal{R}_{\gamma}$ is exactly the regularizer
defined in Sec.~\ref{Optimization}, with the same coefficient
and reduction. It introduces a shrinkage preference on the
modulation factors. Within the active set used by local density
control, primitives with $\gamma_i<0.01$ are removed.

Since the effective scales are $\gamma_iS_i$, a penalty or
threshold on $\gamma_i$ acts on the chosen parameterization
rather than directly constraining Gaussian volume.
Indeed, compensating changes in the base scales can preserve
the effective covariance. The factor therefore serves as an
optimization and primitive-control variable.

\paragraph{Role of the reparameterization.}
The base scales remain learnable, so the modulation does not
enlarge the static family of representable Gaussian
covariances. Its contribution lies in the optimization
parameterization and its interaction with density control.
The reported ablations show improved reconstruction quality
and efficiency when this component is included in the full
framework.

\subsection{Frequency-Aware Mask Generation}
\label{sec:hf_mask}

Given an input image, we first average its color channels to
obtain a grayscale image and downsample it by a factor of
$s=0.5$. We convolve the downsampled image with two Gaussian
kernels having standard deviations $\sigma_1$ and
$\sigma_2=2\sigma_1$, then take the absolute difference of
the filtered images. Each discrete Gaussian kernel is
normalized to unit sum, and zero padding is used to preserve
the spatial dimensions.

The bandwidth is controlled by the schedule parameter
$f\in[0,100]$, which increases linearly with the training
iteration:
\begin{equation}
\sigma_1=
\begin{cases}
0.1+(100-f)\times0.1, & \text{if } f\geq50,\\
0.1+f\times0.1, & \text{otherwise}.
\end{cases}
\label{eq:sigma1}
\end{equation}
The bandwidth increases during the first stage and decreases
during the second stage.

We upsample the absolute response to the original resolution
and apply min--max normalization separately to each image.
The minima and maxima are taken over spatial pixels,
independently of the other images in the batch.
For $f<50$, thresholding the complemented response selects
locations with relatively weak responses at the current scale.
For $f\geq50$, thresholding the direct response selects
strong-response locations, with progressively narrower
filters emphasizing finer structures.
The resulting binary mask, denoted consistently by
$M_{\mathrm{freq}}$, is used in both importance scoring
and the mask-weighted reconstruction loss.

Algorithm~\ref{alg:hf_mask} summarizes the procedure.
The mask provides a stage-dependent spatial selection signal
derived from image-frequency responses. A spatially constant
response has $H_{\mathrm{norm}}=0$ under the stated
stabilization rule, yielding an all-one mask for $f<50$ and
an all-zero mask for $f\geq50$.
Its participation in density control is determined by
the operation sets in Sec.~\ref{sec:density_control_details};
the mask is also used in the reconstruction objective
throughout training.

\begin{algorithm}[t]
\caption{Frequency-Aware Mask Generation}
\label{alg:hf_mask}
\begin{algorithmic}[1]

\Require Input image batch
$I\in
\mathbb{R}^{B\times C\times H_{\mathrm{img}}
\times W_{\mathrm{img}}}$,
schedule parameter $f\in[0,100]$, and scale factor $s$
(e.g., $s=0.5$), where $B$, $C$, $H_{\mathrm{img}}$,
and $W_{\mathrm{img}}$ denote the batch size, number of
channels, image height, and image width, respectively

\Ensure Binary mask
$M_{\mathrm{freq}}\in
\{0,1\}^{B\times1\times H_{\mathrm{img}}
\times W_{\mathrm{img}}}$

\State Convert $I$ to grayscale:
\[
I_{\mathrm{gray}}
\gets
\operatorname{mean}
(I,\mathrm{dim}=1,\mathrm{keepdim}=\mathrm{true}),
\qquad
I_{\mathrm{gray}}
\in
\mathbb{R}^{B\times1\times H_{\mathrm{img}}
\times W_{\mathrm{img}}}
\]

\State Downsample the grayscale image:
\[
I_{\mathrm{down}}
\gets
\operatorname{BilinearInterp}(I_{\mathrm{gray}},s)
\]

\If{$f\geq50$}
    \State $\sigma_1\gets0.1+(100-f)\times0.1$
\Else
    \State $\sigma_1\gets0.1+f\times0.1$
\EndIf

\State $\sigma_2\gets2\sigma_1$

\State Compute the Gaussian kernel sizes:
\[
k_1\gets2\left\lceil3\sigma_1\right\rceil+1,
\qquad
k_2\gets2\left\lceil3\sigma_2\right\rceil+1
\]

\State Generate Gaussian kernels:
\[
\mathcal{K}_1
\gets
\operatorname{GaussianKernel}(k_1,\sigma_1),
\qquad
\mathcal{K}_2
\gets
\operatorname{GaussianKernel}(k_2,\sigma_2)
\]

\State Apply dimension-preserving zero-padded convolution:
\[
I_1^{\mathrm{blur}}
\gets
\mathcal{K}_1*I_{\mathrm{down}},
\qquad
I_2^{\mathrm{blur}}
\gets
\mathcal{K}_2*I_{\mathrm{down}}
\]

\State Compute the absolute difference-of-Gaussians response:
\[
H_{\mathrm{freq}}
\gets
\left|
I_1^{\mathrm{blur}}
-
I_2^{\mathrm{blur}}
\right|
\]

\State Upsample the response:
\[
H_{\mathrm{up}}
\gets
\operatorname{BilinearInterp}
\left(
H_{\mathrm{freq}},
(H_{\mathrm{img}},W_{\mathrm{img}})
\right)
\]

\State Normalize each image independently over its spatial pixels,
where $\varepsilon_{\mathrm{norm}}>0$ is a numerical
stabilization constant:
\[
H_{\mathrm{norm}}
\gets
\frac{
H_{\mathrm{up}}-\min(H_{\mathrm{up}})
}{
\max(H_{\mathrm{up}})-\min(H_{\mathrm{up}})
+\varepsilon_{\mathrm{norm}}
}
\]

\If{$f\geq50$}
    \State $M_{\mathrm{freq}}
    \gets
    \mathbb{I}
    \!\left[H_{\mathrm{norm}}\geq0.5\right]$
\Else
    \State $M_{\mathrm{freq}}
    \gets
    \mathbb{I}
    \!\left[1-H_{\mathrm{norm}}\geq0.5\right]$
\EndIf

\State \Return $M_{\mathrm{freq}}$

\end{algorithmic}
\end{algorithm}

Here, $*$ denotes 2D convolution and
$\mathbb{I}[\cdot]$ denotes the indicator function.

\subsection{Density-Control Definitions and Execution Details}
\label{sec:density_control_details}

This subsection specifies the normalization, counting
support, schedules, and sampling procedure used in
Secs.~\ref{sec:freq_importance}
and~\ref{sec:local_densification_pruning}.
We use $G_t$, $A_t$, $V_i$, $s_i^d(t)$, $s_i^p$,
$H_t$, $P_t$, $b_t$, $\tau_d$, $\tau_p$, and $\tau_s$
following the definitions in
Sec.~\ref{sec:local_densification_pruning}.
Let $t\in\{1,\ldots,T_{\mathrm{total}}\}$ denote the current
training iteration, where $T_{\mathrm{total}}$ is the total
number of training iterations.

\paragraph{Normalization.}
For a nonempty finite index set $\mathcal{S}$, a scalar-valued
collection $\{x_a\}_{a\in\mathcal{S}}$, and
$a\in\mathcal{S}$, define
\begin{equation}
\mathcal{N}_{\mathcal{S}}[x](a)
=
\begin{cases}
\dfrac{x_a-x_{\min}}{x_{\max}-x_{\min}},
& x_{\max}>x_{\min},\\[4pt]
0, & x_{\max}=x_{\min},
\end{cases}
\qquad
x_{\min}=\min_{b\in\mathcal{S}}x_b,
\quad
x_{\max}=\max_{b\in\mathcal{S}}x_b.
\end{equation}
Pixel errors are normalized separately for each image,
whereas pruning statistics are normalized over the current
active set. Constant inputs receive no relative priority.
The unnormalized pixel errors are used for photometric
losses. Density control is skipped when the active set
is empty.

\paragraph{Pixel counting support.}
Let $B_i^j$ denote the image pixels in tiles assigned to
Gaussian $i$ by Compact Box. Let $\alpha_i^j(\mathbf{u})$
be its per-pixel compositing opacity and
$T_i^j(\mathbf{u})$ the transmittance immediately before
processing it in front-to-back order. We define
\begin{equation}
\Omega_i^j
=
\left\{
\mathbf{u}\in B_i^j:
\begin{array}{l}
\text{the compositing loop reaches Gaussian }i,\\
\alpha_i^j(\mathbf{u})\geq\varepsilon_\alpha,\\
T_i^j(\mathbf{u})
\bigl(1-\alpha_i^j(\mathbf{u})\bigr)
\geq\varepsilon_T
\end{array}
\right\},
\end{equation}
where $\varepsilon_\alpha>0$ and
$0<\varepsilon_T<1$ are the renderer's opacity and
early-termination tolerances.
Thus, a pixel is counted only when Gaussian $i$ is reached
by the compositing loop, passes the opacity test, and is
processed before the renderer's early-termination condition
is triggered. We set $\Omega_i^j=\varnothing$ for
$j\notin V_i$.

\paragraph{Densification weight.}
The temporal weight in Eq.~\eqref{eq:densification_score} is
\begin{equation}
u_d(t)
=
\min\!\left\{
1,
\max\!\left\{
0,
\frac{t-t_d^{\mathrm{start}}}
{t_d^{\mathrm{end}}-t_d^{\mathrm{start}}}
\right\}
\right\},
\qquad
\omega(t)
=
\omega_{\mathrm{start}}
+
(\omega_{\mathrm{end}}-\omega_{\mathrm{start}})
u_d(t),
\end{equation}
with
$0<\omega_{\mathrm{start}}\leq\omega_{\mathrm{end}}$
and
$t_d^{\mathrm{start}}<t_d^{\mathrm{end}}$.
For a fixed count and threshold, increasing $\omega(t)$
makes the densification criterion easier to satisfy.
The weight is used only at scheduled densification steps.

\paragraph{Operation schedules.}
Let $t_d^{\mathrm{start}}<t_d^{\mathrm{end}}$ and
$t_p^{\mathrm{start}}\leq t_p^{\mathrm{end}}$ be integer
endpoints within $[1,T_{\mathrm{total}}]$.
For positive integer intervals $\Delta_d$ and $\Delta_p$,
define
\begin{equation}
\begin{aligned}
\mathcal{I}_d
&=
\left\{
t\in\mathbb{Z}:
t_d^{\mathrm{start}}\leq t\leq t_d^{\mathrm{end}},
\quad
(t-t_d^{\mathrm{start}})\bmod\Delta_d=0
\right\},\\
\mathcal{I}_p
&=
\left\{
t\in\mathbb{Z}:
t_p^{\mathrm{start}}\leq t\leq t_p^{\mathrm{end}},
\quad
(t-t_p^{\mathrm{start}})\bmod\Delta_p=0
\right\}.
\end{aligned}
\end{equation}
Density control occurs at
$t\in\mathcal{I}_d\cup\mathcal{I}_p$.
At each such update, we sample $K$ distinct views uniformly
without replacement, where $N_{\mathrm{train}}$ denotes the
number of training views and
$1\leq K\leq N_{\mathrm{train}}$.

The frequency schedule
$f(t)=100t/T_{\mathrm{total}}$ is not restarted for density
control. For either operation $o\in\{d,p\}$, the complemented
and direct masks participate at
\begin{equation}
\mathcal{I}_o
\cap
\{t:t<T_{\mathrm{total}}/2\},
\qquad
\mathcal{I}_o
\cap
\{t:t\geq T_{\mathrm{total}}/2\},
\end{equation}
respectively. Either intersection may be empty.
Mask-weighted supervision follows the same frequency
schedule throughout training, including iterations without
density updates.

\paragraph{Weighted pruning without replacement.}
For $t\in\mathcal{I}_p$, form $H_t$, $P_t$, and $b_t$ as
defined in Sec.~\ref{sec:local_densification_pruning}.
Initialize $U_0=P_t$. For
$\nu=1,\ldots,b_t$, sample
\begin{equation}
\Pr(i_\nu=i\mid U_{\nu-1})
=
\frac{s_i^p}
{\sum_{k\in U_{\nu-1}}s_k^p},
\qquad
i\in U_{\nu-1},
\end{equation}
and update
\begin{equation}
U_\nu
=
U_{\nu-1}\setminus\{i_\nu\}.
\end{equation}
All candidate weights are positive because
$s_i^p>\tau_p\geq0$.
The removal set is
\begin{equation}
R_t
=
H_t\cup\{i_1,\ldots,i_{b_t}\}.
\end{equation}
When $b_t=0$, the sampled subset is empty.
For $t\notin\mathcal{I}_p$, set $R_t=\varnothing$.

\paragraph{Densification and update order.}
Visibility sets and scores are computed from the same
pre-update representation $G_t$.
Pruning, when scheduled, is applied first.
For $t\in\mathcal{I}_d$, densification candidates are
\begin{equation}
D_t
=
\left\{
i\in A_t\setminus R_t:
s_i^d(t)>\tau_d
\right\},
\qquad
\tau_d\geq0.
\end{equation}
Let
\begin{equation}
s_{\max,i}
=
\max\{s_i^x,s_i^y,s_i^z\}.
\end{equation}
For $s_{\max,i}<\tau_s$, with $\tau_s>0$, we clone the
primitive and shift the clone along the positional gradient.
Otherwise, we replace the parent with two smaller Gaussians.
The branch uses base scales, whereas rendering uses
the effective scales $\gamma_iS_i$.

New primitives inherit the parent's modulation parameter
$\beta_i$, and hence $\gamma_i$; their other attributes
follow the underlying clone/split updates.
They receive fresh visibility estimates and scores at
the next scheduled density update and are not pruned
during their creation step.
Every density modification is restricted to $A_t$.

\section{Limitations}
\label{sec:limitations}

Our evaluation focuses on static-scene novel view synthesis
across bounded and urban-scale benchmarks.
The effectiveness of EffGS for time-varying scene
reconstruction remains unevaluated.
The reported results characterize the quality--efficiency
trade-off under the evaluated training configurations.
Although matched-primitive-budget comparisons control
Gaussian counts, they do not characterize performance
across all computational budgets or hardware settings.
Further evaluation is needed to establish how these
trade-offs extend beyond the tested configurations.

\section{Broader Social Impacts}
\label{sec:impact}
This work introduces EffGS, an efficient framework for high-fidelity large-scale 3D scene reconstruction. While our approach positively contributes to environmental sustainability by significantly reducing computational overhead and enables beneficial applications in urban planning and simulation, we acknowledge the potential for misuse in privacy-sensitive contexts, such as unauthorized spatial reconstruction. To mitigate such risks, the deployment of our framework should be governed by clear ethical guidelines, rigorous dataset sanitization prior to training, and regulatory oversight.

\end{document}